\documentclass[10pt]{article}

\usepackage{amsmath, amsfonts, amssymb, mathrsfs, amsthm}
\usepackage{lineno}
\usepackage{hyperref}
\usepackage{stix}

\usepackage{mathtools} 
\usepackage{extarrows}

\modulolinenumbers[5]

\usepackage{wrapfig}

\def\drawplusplus#1#2#3{\hbox to 0pt{\hbox to #1{\hfill\vrule height #3 depth
      0pt width #2\hfill\vrule height #3 depth 0pt width #2\hfill
      }}\vbox to #3{\vfill\hrule height #2 depth 0pt width
      #1 \vfill}}

\usepackage[all]{xy}
\usepackage{graphicx}
\newtheorem{thm}{Theorem}
\newtheorem{mydef}{Definition}

\newtheorem*{thm* }{Theorem}
\newtheorem*{mydef*}{Definition}
\newtheorem*{mylemma*}{Lemma}
\newtheorem*{myconjecture* }{Conjecture}

\begin{document}


\title{The General Theory  of General Intelligence: \\
A Pragmatic Patternist Perspective}

\author{Ben Goertzel}





\maketitle

\begin{abstract}
A multi-decade exploration into the theoretical foundations of artificial and natural general intelligence, which has been expressed in a series of books and papers and used to guide a series of practical and research-prototype software systems, is reviewed at a moderate level of detail.  The review covers underlying philosophies (patternist philosophy of mind, foundational phenomenological and logical ontology), formalizations of the concept of intelligence, and a proposed high level architecture for AGI systems partly driven by these formalizations and philosophies.    The implementation of specific cognitive processes such as logical reasoning, program learning, clustering and attention allocation in the context and language of this high level architecture is considered, as is the importance of a common (e.g. typed metagraph based) knowledge representation for enabling "cognitive synergy" between the various processes.   The specifics of human-like cognitive architecture are presented as manifestations of these general principles, and key aspects of machine consciousness and machine ethics are also treated in this context.   Lessons for practical implementation of advanced AGI in frameworks such as OpenCog Hyperon are briefly considered.
\end{abstract}



\tableofcontents

\section{Introduction}

The relation between formal theory, conceptual theory and experimentation in AI has historically been subtle and dialectical, as in many disciplines where engineering is allied with frontier science.  As a few examples:

\begin{itemize}
\item Genetic algorithms were a case where strong conceptual analogies to biology led to robust experimentation, which was followed only significantly later by useful formal theoretical understanding came only later \cite{Goldberg_Design}.   
\item Deep neural nets were an example where weak analogies to biology were followed by fairly useful formal theory (e.g. regarding hierarchical neural approaches to function approximation and reinforcement learning), which then  for decades led only to toy-scale and relatively unimpressive practical examples, until supporting technologies matured enough that the real-world power of the ideas could be realized experimentally \cite{mccorduck2004machines}.  
\item Logic-based AI has been strong on theory for quite some time, and there is an increasing suspicion that it's going to finally come into its practical prime over the next 5 years with the rise of neural-symbolic systems \cite{garcez2020neurosymbolic}.   Modern work on ML-based guidance of theorem-proving combines empirical experimentation with formal theory in a fascinatingly intricate way (e.g. \cite{DBLP:conf/mkm/UrbanJ20}).
\end{itemize}

Artificial General Intelligence (AGI) research, considered as a subset of AI research, has also combined theory and experimentation in various and complex ways.  At the one extreme, there has been the approach of starting with a general theory of AGI and then deriving practical systems from this theory and implementing them.   Marcus Hutter and his students have been the best example of this approach, with Hutter's Universal AI theory \cite{Hutter2005} serving as a credible (though debatable in many respects) general theoretical AGI approach and a number of relatively practical proto-AGI systems emerging from it \cite{Everitt2016}.  Arthur Franz's work has perhaps gone the furthest toward building a practical bridge between Hutter's universal AGI theory and the realm of practically usable AGI systems \cite{franz2019william} \cite{franz2015artificial} \cite{franz2021theory}.   

At the other extreme, there is the currently more common approach of working toward AGI by creating more and more powerful practical ML and RL systems, experimenting with them and seeing what they can do, and then working out theoretical explanations of observed AGI system behaviors as needed.   The various attempts underway to work toward AGI by creating more and more powerful neural net architectures, incorporating e.g. deep and reinforcement learning networks combined and end-to-end trained using backpropagation, are primarily in this spirit.  There is a broad underlying conceptual framework, a rather loose analogy to aspects of human neuroscience, and a fairly robust set of relevant mathematical tools, but there is not much of an attempt to derive the details of an AGI architecture from an overall conception of what a mind is.

My own approach to AGI over the last several decades has been on the whole more theoretically than experimentally driven -- with an integrative "cognitive systems theory" approach including mathematics along with other disciplinary influences, rather than a primarily mathematical approach a la Hutter.  However, I have also been involved with a series of projects aimed at implementing practical software systems according to these ideas -- starting with the Webmind system (1997-2001) \cite{BabyWebmind}, then the Novamente Cognition Engine (2001-2008) \cite{goertzel2007novamente}, then OpenCog (2008-2021) \cite{Goertzel2008a} and now the new OpenCog Hyperon version \cite{Hyperon2021}.   Each of these systems has been used behind some practical narrow-AI applications, and has also been used for numerous AGI-prototyping experiments aimed more at building understanding of various aspects of the AGI problem that at achieving impressive practical results.   The comprehensive theoretical framework and high level design for AGI presented by myself, Cassio Pennachin and Nil Geisweiller in our 900-page 2014 work {\it Engineering General Intelligence} \cite{EGI1} \cite{EGI2} built on my earlier theoretical works such as \cite{Goertzel2006a} \cite{Goertzel1994} \cite{Goertzel1993a} \cite{Goertzel1993} \cite{Goertzel1997}\cite{Goertzel2001}, but also on the many practical lessons learned from our experimentation with these systems.

In this paper I summarize and relatively concisely review key aspects of the long series of explorations I have made over the last few decades into the theoretical underpinnings of general intelligence -- substantially focused on engineered AGIs, but largely intended as applicable more broadly to natural general intelligences (such as humans, other animals, or currently unknown-to-us forms of natural general intelligence) as well.   

The structure of the review is first of all from the general and abstract to the precise and engineering-oriented.   That is, I begin with a conceptual and mathematical vision of "what general intelligence is", and then proceed to introduce a series of conceptual and mathematical simplifications, approximations and assumptions that leads in the direction of practically implementable AGI designs and systems.   The thrust is to begin with "general intelligence in general" and then arrive at key elements of the OpenCog Hyperon design as a specialization of general principles of general intelligence -- with an understanding that it's the journey as well as the origin and destination that's of interest here.  

Following this voyage from the general to the particular, in Sections \ref{sec:cons} and \ref{sec:ethics} I back up to the general again, considering questions of consciousness and ethics which pertain to the consilience of AGI designs and principles with the broader context of humanity and the universe at large.   

The paper may be considered as something similar to a carefully structured annotated bibliography of many of my prior works on AGI theory -- reasonably thorough references to these prior publications are given, and the discussion here is more oriented toward capsulizing the most striking high-level conclusions rather than trying to convey all the arguments, equations, examples and particulars.   As with any body of complex math, science or engineering concepts, if you really want to understand you'll have to follow the references and put in the time to absorb the details.

The intellectual and practical quests summarized here are by no means complete -- neither I nor anybody else on this planet has yet built an AGI with capability at the human level or beyond; and nor has anyone here yet articulated a comprehensive theory of general intelligence that can be used to guide AGI design in the precise and careful way that, say, fluid dynamics and aerodynamic theory can be used to guide flying-machine design.   However, I do believe the theoretical developments summarized here constitute significant progress toward a useful general theory of general intelligence; and my strong hypothesis is that following the guidance of these theoretical ideas in the implementation domain comprises a highly viable approach to realizing powerful AGI.  A great deal has been learned in preceding decades via exploring multiple iterations of the theoretical concepts given here and by building and running practical systems inspired by various aspects of these concepts; and in my view all this learning, put together with today's unprecedentedly powerful compute fabrics and voluminous data sets and streams, creates an outstanding condition for multidimensional accelerated progress moving forward.

\subsection{Summary of Key Points}

Given the somewhat immense scope of the subject matter, it may be useful to give a relatively compacted run-through of the main issues to be touched:

\begin{enumerate}
\item The "patternist philosophy of mind", in which the aspects of intelligence most relevant from an engineering perspective are viewed in terms of the understanding of a mind as the set of patterns associated with an intelligent system
\item General aspects of intelligent function like evolution and self-organization, and aspects of cognitive network structure and dynamics, are conceived in a patternist way
\item A formalization of the concept of "pattern", grounding pattern in a formal theory of complexity/simplicity that embraces algorithmic information theory but also frames the concepts more generally in terms of "combination systems" of simple elements that combine to produce other elements in the manner of an abstract algorithmic chemistry
\item G. Spencer Brown's {\it Laws of Form} and related thinking regarding "distinction graphs" is introduced as a more foundational ontological and phenomenological layer within which the formalization of pattern, simplicity, combination, function application, process execution and related concepts can be situated
\item Distinction graphs are seen to naturally extend into distinction metagraphs, with typed nodes and links including e.g. types related to temporal relationships.   These metagraphs can be taken as a foundational knowledge representation and meta-representation scheme for AGI theory and practice.
\item Paraconsistent, probabilistic and fuzzy logic can be grounded naturally in distinction metagraphs and their symmetries and emergent properties
\item Execution and analysis of programs in appropriate languages can be grounded in distinction metagraphs via Curry-Howard correspondences between these languages and logics that are grounded in distinction metagraphs
\item Intelligence in general must be considered as an open-ended phenomenon without any single scalar or vectorial quantification.  However, intelligent systems can be quantified in multiple respects, including e.g. joy, growth and choice, and also including goal-achievement skill.
\item Formalization of the "goal-achievement skill" aspect of intelligence in terms of algorithmic information theory is interesting in multiple respects, including the simple formal models of extraordinarily intelligent though physically infeasible agents (e.g. AIXI$^{tl}$ and the Godel Machine) that it naturally corresponds to
\item The activity of these impractical formal extraordinarily intelligent agents can be associated with formal models of the world constructed according to elegant information-theoretic principles like "Maximal Algorithmic Caliber"
\item Achievement of reasonably high degrees of general intelligence under conditions of constrained resources relies heavily on "cognitive synergy" -- the property via which different sorts of learning processes associated with different kinds of practically relevant knowledge are able to share intermediate internal state and help each other out of learning dead-ends and bottlenecks
\item Approximation of impractical formal models of extraordinarily intelligent agents in terms of practically achievable Discrete Decision Systems (DDSs) seeking incremental reward maximization via sampling and inference guided action selection is a worthwhile approach to practical AGI design.   These DDSs can often be executed in terms analyzable as greedy algorithms or approximate stochastic dynamic programming.
\item Combinatory Function Optimization (COFO) systems -- which seek to maximize functions via guiding function-evaluation using sampling and inference guided selection of combinations within a combination system -- are introduced as a species of DDS particularly useful within AGI architectures.  
\item Practical cognitive systems are viewed as recursive DDSs aimed at carrying out organismic goals (like pursuing joy, growth, choice, survival, discovery of new things, etc.), via choosing actions via methods that rely on COFO systems oriented toward various function-optimization subgoals.
\item Key practical cognitive algorithms like probabilistic logical inference, evolutionary and probabilistic program learning, agglomerative clustering, greedy pattern mining and activation spreading based attention allocation (used e.g. in the OpenCog AGI design) are represented as COFO systems.   
\item The formalization of these key cognitive algorithms in COFO terms is driven by the representation of e.g. logical inference rules, program execution steps and clustering steps as operations within, upon and by distinction metagraphs.  This common representation is critical for practical achievement of cognitive synergy.
\item Practical COFO systems implementing these key cognitive algorithms can be approximatively represented using Galois connections, which -- as shown by theorems summarized here -- allows them to be approximatively implemented in software via chronomorphisms (folds and unfolds) over typed metagraphs.   
\item Algebraic associativity properties of combinatory operations (as represented by edges in typed metagraphs interpreted as programmatic metagraph transformations) play a key role in enabling practical general intelligence given realistic resource constraints.   Cost-associativity of combinatory operations underlying cognition is critical for construction of subpattern hierarchies (hierarchical knowledge representation), whereas associativity of combinatory operations underlying COFO representations of cognitive processes is critical for mapping these COFO dynamics into chronomorphisms.
\item The cognitive architecture of human-like intelligences, as articulated via various theories and researches within the cognitive science discipline (and illustrated here in a series of cognitive architecture diagrams), can be viewed as a way of arranging these key cognitive algorithms in an overall DDS configured to operate within the sorts of resource constraints characterizing human brains and bodies
\item Essential properties of AGI knowledge representations and programming languages can be derived from these considerations -- this is part of the design process currently being undertaken regarding OpenCog Hyperon.
\item "Consciousness" in AGI systems may be understood as a holistic phenomenon characterized by a number of different properties; human-like consciousness is a particular manifestation of general consciousness which is driven by key properties of human-like cognitive architecture including cognitive synergy and attention-focusing.
\item Ethics in AGI systems will take different manifestations as these systems mature in their cognitive capabilities; advanced self-reflecting and self-modifying AGI systems, if appropriately designed and educated, should be able to achieve a level of "reflective ethics" beyond what is possible within human brain/mind architecture.   
\item Achieving advanced reflective ethics will require the right cognitive architecture (e.g. the GOLEM framework) but also the right situations and interactions during the system's growth phase, e.g. focus on broadly beneficial goals rather than narrow goals primarily benefiting particular parties.
\end{enumerate}

\section{Patternist Philosophy of Mind}

The relation between AGI as a practical endeavor (aimed at building and teaching and deploying systems) and {\it philosophy-of-mind}   -- as distinct from scientific psychology -- is not entirely obvious.   Scientific psychology is driven fundamentally by empirical data (regarding human, animals and sometimes computer models or AI systems), and seeks to form theories that explain this data.   Philosophy of mind is driven fundamentally by conceptual reflection, though it may incorporate empirical results into this reflection.

The strongest argument for including philosophy of mind foundationally in one's path to AGI is that, by its nature, the quest for {\it general} intelligence goes beyond the particular intelligent systems one currently has direct evidence on.   And, furthermore, the available data about general intelligence is very scant relative to the complexity of the phenomenon, meaning that extrapolating from this data is likely to yield theories that focus too much on the specific aspects of intelligence that happen to have been most studied so far, rather than coming to grips effectively with the overall nature of intelligence.   An example of this latter phenomenon would be the outsized influence of models of the mammalian visual system on contemporary cognitive science and AI design.   The cognitive neuroscience of vision is especially well developed because vision experiments are relatively easy to run on monkeys and other mammals, and it's partly due to this that our currently best developed neural net architectures are so markedly hierarchical, mirroring the coarse structure of visual cortex (and much less effectively mirroring the structure of other parts of the cortex that more richly mix up hierarchical and combinatory connections \cite{Lynch1986}).

This gels well with the argument that if one's fundamental understanding of {\it what mind is} is too weak, one may fail at AGI due to screwing around in dead ends that would have been ruled out by a deeper conceptual understanding.   As a rough analogy, while modern biology doesn't include a crisp ironclad definition of "what life is," there's no doubt that the modern conceptual understanding of the nature of organismic life and its relation to chemistry below and ecology above has been extremely critical to recent practical progress in bringing evolutionary biology beyond simplistic Neo-Darwinism \cite{Noble7} -- and that fleshing out this conceptual understanding further will be critical for ongoing biological revolutions like achieving radical human life extension via a combination of molecular and systems biology \cite{Goertzel2014AGIRev}.   

On the other hand, the obvious argument {\it against} paying attention to philosophy-of-mind in an AGI engineering context would be that, for instance, solid-state physics has created all sorts of amazing new forms of matter without fundamentally resolving the nature of {\it what matter is} -- the latter being a topic confusingly wrapped up with interpretations of quantum measurement and diverging speculative theories of unified physics.   Philosophy tends to create conceptual tangles whereas practical engineering and experimentation tends to cut through confusion, along the way clarifying which thorny intellectual messes actually need to be untangled and which can be shoved off to the side while real work proceeds.

As you may guess I have sought a middle path of sorts.  In {\it The Hidden Pattern} \cite{Goertzel2006a}, I have outlined in detail the philosophy of mind underlying my own work on AGI.   As I elaborate there, my view is that philosophy of. mind provides a valuable starting-point for practical AGI design -- but also has its limits.   One reaches a point where philosophy doesn't provide adequate help with the decisions at hand.   Part of the modus operandi of the technical theoretical work summarized in this article is to use mathematics as a bridge between philosophy and engineering.  There are still of course gaps at either end, and leaps to be made to get from the philosophy to the math and from the math to the engineering.  But these leaps are smaller than if one tries to get from philosophy to engineering directly.

\subsection{Patternist Principles} \footnote{Some of the text in this section is adapted from various parts of {\it Engineering General Intelligence, Vol. 1} \cite{EGI1}}

 {\it The Hidden Pattern} outlines what I call a "patternist philosophy of mind" --
  a general approach to thinking about intelligent
 systems, based on the very simple premise that mind is made of pattern.  I.e.
 that a mind is a system for recognizing patterns in itself and the world, critically
including patterns regarding which procedures are likely to lead to the achievement
 of which goals in which contexts.

In patternism the mind of an intelligent system is conceived as the (fuzzy)
set of patterns in that system, and the set of patterns emergent between that system
and other systems with which it interacts. The latter clause means that the patternist
perspective is inclusive of notions of distributed intelligence \cite{Hutchins1995}. Basically, the
 mind of a system is the fuzzy set of different simplifying representations of that
system -- as presented in various contexts -- that may be adopted.

Intelligence may be partially conceived, in this framework, as the ability to achieve
 complex goals in complex environments; where complexity itself may be defined as
 the possession of a rich variety of patterns. A mind is thus a collection of patterns
 that is associated with a persistent dynamical process that achieves highly-patterned
 goals in highly-patterned environments.

An additional hypothesis made within the patternist philosophy of mind is that
reflection is critical to intelligence. This lets us conceive an intelligent system as a
dynamical system that recognizes patterns in its environment and itself, as part of its
 quest to achieve complex goals.

While this approach is quite general, it is not vacuous; it gives a particular structure
 to the tasks of analyzing and synthesizing intelligent systems. About any would-be
intelligent system, we are led to ask questions such as:

\begin{itemize}
\item How are patterns represented in the system? That is, how does the underlying infrastructure of the system give rise to the displaying of a particular pattern in the system's behavior?
\item What kinds of patterns are most compactly represented within the system?
\item What kinds of patterns are most simply learned?
\item What learning processes are utilized for recognizing patterns?
\item What mechanisms are used to give the system the ability to introspect (so that it can recognize patterns in itself)?
\end{itemize}

Addressing these questions leads to the identification of a few key dynamics as driving real-world intelligent systems, e.g.

\begin{itemize}
\item {\bf Evolution} -- conceived as a general process via which patterns within a large pop- ulation thereof are differentially selected and used as the basis for formation of new patterns, based on some "fitness function" that is generally tied to the goals of the agent.
\item  {\bf Autopoiesis} -- The process by which a system of interrelated patterns maintains its integrity, via a dynamic in which whenever one of the patterns in the system begins to decrease in intensity, some of the other patterns increase their intensity in a manner that causes the troubled pattern to increase in intensity again.
\item  {\bf Association} -- Patterns, when given attention, spread some of this attention to other
patterns that they have previously been associated with in some way. Furthermore, there is Peirce?s law of mind [Pei34], which could be paraphrased in modern terms as stating that the mind is an associative memory network, whose dynamics dictate that every idea in the memory is an active agent, continually acting on those ideas with which the memory associates it.
\item  {\bf Pattern creation} -- Patterns that have been valuable for goal-achievement are
mutated and combined with each other to yield new patterns.
\item   {\bf Hierarchical network} -- Patterns are habitually in relations of control over other
patterns that represent more specialized aspects of themselves.
\item  {\bf Heterarchical network} -- The system retains a memory of which patterns have pre-viously been associated with each other in any way.
\item {\bf Dual network} -- Hierarchical and heterarchical structures are combined, with the dynamics of the two structures working together harmoniously. Among many possible ways to hierarchically organize a set of patterns, the one used should be one that causes hierarchically nearby patterns to have many meaningful heterarchical connections; and of course, there should be a tendency to search for heterarchical connections among hierarchically nearby patterns.
\item {\bf Self structure} --  A portion of the network of patterns forms into an approximate image of the overall network of patterns.
\end{itemize}

If the patternist philosophy of mind is a useful one, then the success of any AGI design or system will depend largely on whether these high-level structures and dynamics can  be made to emerge from the synergetic interaction of the given representation and
algorithms, when they are utilized to control an appropriate agent in an appropriate
environment.

\subsection{Cognitive Synergy}

An important elaboration of the basic patternist philosophy of mind is the notion of "cognitive synergy."

Cognitive synergy begins with the observation that, with respect to certain classes of goals and environments -- such as those with which humans are generally concerned -- an intelligent system operating within feasibly limited computational resources requires a ``multi-memory'' architecture, meaning the possession of a number of specialized yet interconnected knowledge types, including: declarative, procedural, attentional, sensory, episodic and intentional (goal-related). These knowledge types may be viewed as different sorts of patterns that a system recognizes in itself and its environment.  Such a system must possess knowledge creation (i.e. pattern recognition / formation) mechanisms corresponding to each of these memory types. These mechanisms are what I refer to as ``cognitive processes.``

The next step is the observation that each of these cognitive processes, to be effective, must have the capability to recognize when it lacks the information to perform effectively on its own; and in this case, to dynamically and interactively draw information from knowledge creation mechanisms dealing with other types of knowledge.  This cross-mechanism interaction must have the result of enabling the knowledge-type-specific knowledge creation mechanisms to perform much more effectively in combination than they would if operated non-interactively.  This is ``cognitive synergy`` -- a conceptual notion which, as pursued in \cite{DBLP:journals/corr/Goertzel17} and noted below, can also be formulated in a rigorous mathematical way by means of category theory.

\section{Foundational Ontology } \label{sec:ontology}

Patternism is a conceptual theory rather than a formal one, and may be turned into a formal theory in various different ways.  Each act of formalization, like all other acts, involves some gain and some loss; one would not want to replace the conceptual theory of patternism with any of its formalizations, but to proceed from the starting-point of patternism toward practical goals like AGI design and engineering, formalization is a natural step.

The formal structures described in this section are presented, proximally, as ways of describing the phenomenology of cognitive systems as experienced from the inside ("first person"), as well as the presence of cognitive systems as experienced by other cognitive system interacting with them ("second person") or observing them in a relatively decoupled way ("third person").   As such they are more along the lines of very abstract theoretical psychology than AGI design per se.   However, in Section \ref{sec:metagraph-agi} it will be pointed out that these same structures can also be taken directly as dynamic data structures underlying AGI systems (e.g. OpenCog Hyperon), creating a pleasantly direct route to AGI systems capable of modeling their own behaviors and experiences.

\subsection{From Laws of Form to Paraconsistent and Probabilistic Logic}

My current favorite avenue for formalizing patternism is to begin by connecting it to another interesting conglomeration of philosophical, mathematical, scientific and engineering considerations -- the {\it Laws of Form} paradigm, initiated by G. Spencer Brown in his book by that name \cite{SpencerBrown1967} and extended and enriched dramatically by Louis Kauffmann and others \cite{KauffmannSS}.

The {\it Laws of Form} paradigm could be thought of as its own sort of ``patternism'' -- or else perhaps as ``distinctionism.''   One starts the analysis and synthesis process with elementary observations, where the understanding is that the most elementary sort of observation is a {\it distinction} -- just an act of distinguishing some stuff from some other stuff.   One can also look at recursively paradoxical distinctions -- distinctions that distinguish themselves from themselves -- which Spencer-Brown refers to as ``imaginary forms``, with closely analogous properties to imaginary numbers.   

Ordered pairs of distinctions (2D distinctions), with the appropriate simple assumptions, can be shown isomorphic to recursively paradoxical distinctions -- a result that turns out interestingly relevant to our current AGI-oriented work with PLN (probabilistic logic networks) in the OpenCog system, by way of connections between paraconsistent and probabilistic logic.  

Roughly, if one considers an unmarked state to be True, and a distinguished state to be False (so that distinction is a form of negation), then a recursively paradoxical state "This state is False" can be resolved in time in two ways

\begin{gather*}
\ldots, True, False, True, \ldots \\
\ldots, False, True, False, \ldots
\end{gather*}

\noindent and one can map these two "real" and two "imaginary" states into four 2D truth values 

\begin{gather*}
(True,True) = \textrm{both true and false}\\
(True, False) = \textrm{true}\\
(False,True) = \textrm{false}\\\
(False, False) = \textrm{neither true nor false}\
\end{gather*}

\noindent. One can articulate the algebra of conjunction, disjunction and negation on these truth values \cite{patterson1998implicit}, thus arriving at a simple paraconsistent logic.  Extending this to account for varying amounts of evidence one obtains uncertain truth values of the form $(w^+, w^-)$ where each component is in $[0,1]$, and $w^+$ and $w^-$ represent respectively the number of situations in which a certain proposition received positive or negative evidence, where it's understood that some situations may contain both positive and negative evidence and some may contain neither.   The algebra of these uncertain paraconsistent truth values can then be shown isomorphic to the PLN algebra of probabilities and weights-of-evidences \cite{goertzel2021paraconsistent}.   That is, PLN Simple Truth Values are of the form $(s,c)$ where $s \in [0,1]$ is a probability value and $c \in [0,1]$ denotes the confidence in that probability value; there is a straightforward rescaling from these STVs into paraconsistent truth values of the form $(w^+, w^-)$.  Probabilistic and paraconsistent logic are thus revealed as different ways of scaling basic counts of the positive and negative evidence contained in observations.

\subsection{From Distinction Graphs to Dynamic Knowledge Metagraphs}

The paper {\it Distinction Graphs and Graphtropy} \cite{Graphtropy} builds on the Laws of Form paradigm by introducing ``distinction graphs'' -- in which a symmetric link is drawn between two observations, relative to a given observer, if the observer cannot distinguish them (basically an "observation" can be considered as "something that can be distinguished").   Graphtropy -- basically the percentage of possible binary distinctions that the graph includes -- is introduced as an extension of logical entropy \cite{ellerman2013introduction} from partitions to distinction graphs.  Conditional graphtropy indicates the amount of additional distinction added by one distinction graph relative to another.   Extensions such as probabilistic and quantum distinction graphs are relatively straightforward, and an analogue of the maximum entropy principle for distinction graphs has been developed.

\begin{figure}[htb]
\centering
\includegraphics[width=10cm]{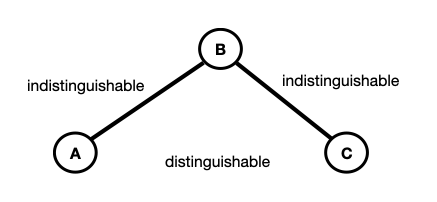}
\caption{Simple distinction graph.   Nodes represent observations; a link between two nodes indicates that, for the observer to whom the graph is relative, these two observations are indistinguishable.  Labels like A, B, C are for the reader's delectation and aren't required as part of the formal distinction graph at this simple level.}
\label{fig:dist-graph}
\end{figure}

Layering additional typed nodes and links atop distinction graphs, one quickly arrives at logical and programmatic representations.   In typical OpenCog notation, a link in a simple (crisp) distinction graph is a SimilarityLink with truth value $(1,1)$, and the absence of a link in a simple distinction graph is a SimilarityLink with truth value $(0,1)$.     Asymmetric distinction links also make sense, where $a \rightarrow b$ would indicate that if an observer had $a$ in mind, then they would not be able to notice $a$ shifting into $b$.

\begin{figure}[htb]
\centering
\includegraphics[width=10cm]{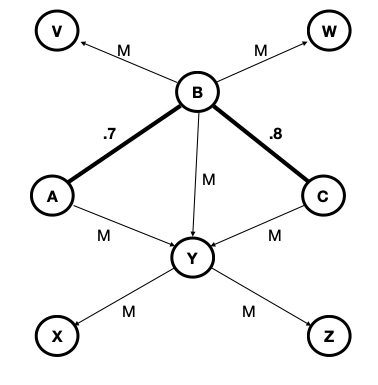}
\caption{Distinction graph enhanced with conceptual groupings (ConceptNodes) and weighted links.   Links labeled "M" denote a membership relationship between a node connoting a group of observations (or a group of observation-groups), and a group or observation being grouped.   Distinctions between nodes connoting groups are naturally labeled with probabilistic or fuzzy weights.}
\label{fig:dist-graph-weighted}
\end{figure}

\begin{figure}[htb]
\centering
\includegraphics[width=12cm]{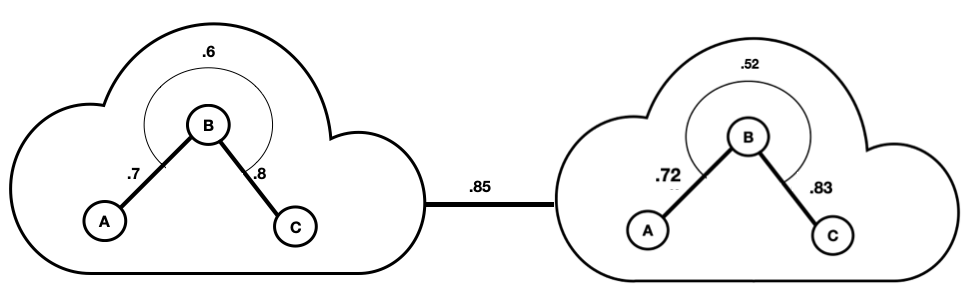}
\caption{Distinction graph enhanced with metagraph features: distinctions between distinctions (links between links) and distinctions between whole distinction graphs (links between subgraphs).}
\label{fig:dist-metagraph}
\end{figure}

One can enhance the distinction graph framework by introducing ConceptNodes that group distinction graph nodes representing elementary observations, with MemberLinks between a ConceptNode and the elementary observation nodes it groups.    One then gets probabilistic symmetric and asymmetric distinctions between these ConceptNodes -- i.e. PLN SimilarityLinks and InheritanceLinks \cite{PLN}.   One also gets an extension from distinction graphs to distinction hypergraphs -- including distinctions between distinctions -- and distinction metagraphs which include distinctions between distinction graphs.    Predicate logic with its abstractions and quantifiers can be considered as a shorthand for elementary uncertain term logic relationships \cite{ikle2010grounding}, so we can build up the full apparatus of logic from the distinction graph infrastructure -- in essence, basically logic emerges as a notational system for describing recursively nested symmetries in distinction graphs. 

\begin{figure}[htb]
\centering
\includegraphics[width=10cm]{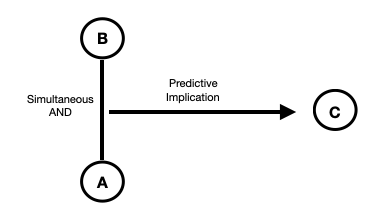}
\caption{Adding links representing temporal relationships to distinction graphs enables numerous representational capabilities, including representation of decision trees that summarize executable functions.}
\label{fig:dist-temporal}
\end{figure}

Considering TemporalConceptNodes that group members co-occurring in a specific interval of time, we can then look at PredictiveImplication links between them.   Introducing also links representing disjunction (XORLink in particular, but typically this is introduced along with ANDLink, ORLink and NOTLink, and temporal versions of these relationships like SequentialAND, SimultaneousOR, etc. \cite{PLN}), we then can represent decision trees or decision dags.   

A typical decision tree can be viewed as partitioning its inputs, where a partition cell contains inputs that all map into the same output value.   Taking a more subjective view of things, one can look at a decision tree relative to a given observer and say that two inputs are distinguished by the tree if they lead to outputs that are distinguishable by the observer.   Or in the context of asymmetric distinction graphs: Input $y$ is distinguishable from the position of input $x$ by the tree, if the output of the tree on $y$ is distinguishable from the position of the output of the tree on $x$.   We can model this by viewing a decision tree as taking inputs consisting of sets of "stars" in a distinction graph (a star consisting of a node plus every other node that it's directly linked to and is hence indistinguishable from), rather than sets of individual distinction graph nodes, and by viewing the output of the decision tree as a set of stars as well.      

Introducing an ExecutionLink that allows specification that a certain node is equivalent to the application of a certain decision dag to a certain input, and introducing nodes that refer to whole decision dags (i.e. "decision dag reflection"), one arrives at compacted decision dags referred to as Combinatorial Decision Dags (CoDDs) \cite{goertzel2020combinatorial}.   CoDDs possess the same abstraction properties as standard SK combinatory logic, and they have the favorable property that if CoDD $f$ extends CoDD $g$ then the former must have higher logical entropy than the latter -- so there is a sort of correlation between complexity as measured via decision-dag size and complexity as measured by counting distinctions.

\begin{figure}[htb]
\centering
\includegraphics[width=10cm]{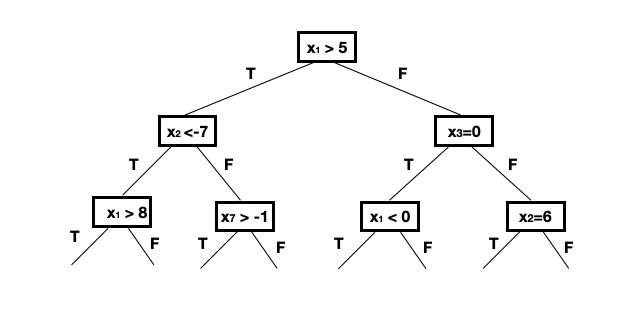}
\caption{Decision Dag, summarizing the execution of a function as a series of simple binary decisions.   This is a highly time-efficient and space-inefficient representation of a function.}
\label{fig:decision-dag}
\end{figure}

\begin{figure}[htb]
\centering
\includegraphics[width=10cm]{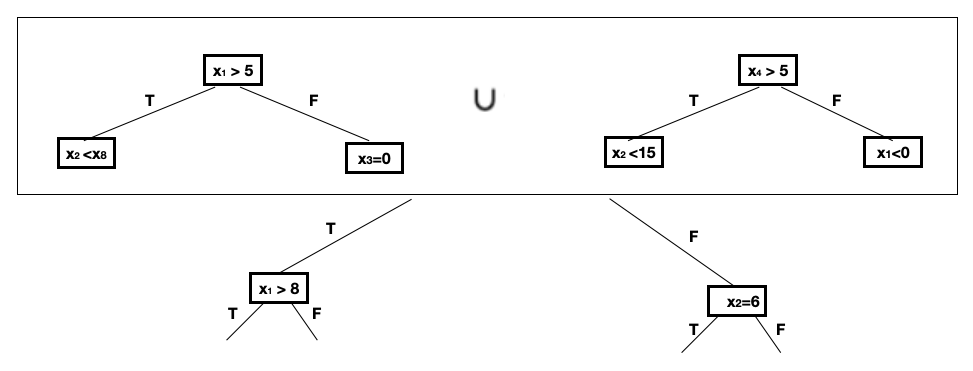}
\caption{Simple example of Combinatorial Decision Dag (CoDD), showing the way a CoDD can encapsulate decisions based on whole decision sub-dags, considering decision sub-dags as inputs.   This is one example of the type of recursion that CoDDs have that ordinary decision dags don't.}
\label{fig:simple-codd}
\end{figure}

In these ways, weighted, typed distinction metagraphs (which can still be considered a form of distinction graph) may be taken as an elementary knowledge-structure, for use in analyzing natural intelligences or other complex systems, and designing and and implementing AGIs and other artificial complex systems.   

\subsubsection{Distinctions Transcending Distinctions}

Considering these constructs in an AI context, it's worth noting that we are operating at a level here at which the "symbolic" versus "subsymbolic" dichotomy that has played such a large role in the history of the AI field \cite{goertzel2014artificial} is made to appear so coarse as to be essentially irrelevant.  The elementary observations defining the distinctions in a distinction graph are "subsymbolic" in the extreme, whether they are distinctions between physical conditions inside a robot's sensor, or distinctions between RAM states in a computer carrying out a mathematical proof.   Networks of patterns built up from these elementary distinctions will embody various forms of semiosis and reference including iconicity, indexicality and symbolism \cite{peirce1991}, and the dynamics of sub-metagraphs interpreted as executable code may embody pattern-recognition algorithms conventionally referred to as "subsymbolic" or inference algorithms conventionally referred to as "symbolic" -- or other sorts of algorithms defying simple labeling in these terms.   

These foundational distinction-based representations also transcend commonly posed dichotomies between localized and distributed representations of knowledge.   Complex patterns of distinctions may exist and have causal influence, whether or not explicitly symbolized in terms of small sets of nodes and links.   Important knowledge for various purposes may also be contained in small number of links or single distinctions.   We are at a meta-representational level where highly localized, broadly distributed or intermediately distributed/localized knowledge representation are all transparently woven from the same fabric.

In Section \ref{sec:hyperon} we will see that, in the context of explicitly metagraph-centric AGI architectures like OpenCog Hyperon, the ability of distinction metagraphs to represent a level more fundamental than typical symbolic vs. subsymbolic or localized vs. distributed considerations manifests itself among other ways in terms of explicitly "neural-symbolic" algorithmics acting on a knowledge metagraph whose meta-representational capabilities encompass neural-network and logical-theorem type representations among others.

\subsection{Measuring Simplicity and Pattern} \label{sec:pattern} 

A key step in creating formalizations inspired by the patternist philosophy of mind is the formalization of the concept of pattern itself.   In early works on patternist models of intelligence, an algorithmic information theory style formalization of pattern was used: basically a pair $(f,x)$ is a pattern in $y$ if 

\begin{itemize}
\item $f*x=y$ where $*$ is an appropriate combinatory operation
\item $\sigma(f) + \sigma(x) < \sigma(y)$ where $\sigma$ is an appropriate simplicity measure (for example $\sigma(x)$ could measure the length of $x$ as expressed in a given language).
\end{itemize}

\noindent More recent work extends and enriches this perspective but with the same fundamental spirit.

In \cite{goertzel2020grounding} a formal theory of simplicity is introduced, in the context of a "combinatory" computation model that views computation as comprising the iterated transformational and compositional activity of a population of agents upon each other.   Conventional measures of simplicity in terms of algorithmic information etc. are shown to be special cases of a broader understanding of the core "symmetry" properties constituting what is defined as a Compositional Simplicity Measure (CoSM).

The combinatory model of computation concerns systems that are composed of a set of elements that act on and transform each other to produce other elements, and join with each other to produce new elements.    This is conceptually very similar to what I have called a "self-generating system" in prior publications \cite{Goertzel1994} \cite{GoertzelHP} \cite{EGI1}, but with updated formal particulars.   Consider a space $\mathcal{E}$ of entities endowed with a set of binary operations $*_i : \mathcal{E} \rightarrow \mathcal{E}^{k_i}, i = 1 \ldots K$.   The operations $*_i$ may be thought of e.g. as reactions via which pairs of entities react to produce sets of entities, or as combinatory operators via which pairs of entities combine to produce sets of new entities.

An entity paired with a combinatory operator, say $x *_2$, can be interpreted as a function acting on entities, and can thus be modeled e.g. as a CoDD.   Or an entity in itself, say $x$, can be interpreted as a function acting on pairs $(*_i, y)$, and thus modeled as a CoDD itself.   In this way a combinatory system can be modeled as a Scott domain of functions that act on other functions in the domain to produce other functions in the domain \cite{gierz2003continuous}, and/or modeled as a system of CoDDs that take other CoDDs as inputs.

The simplicity of an entity can then be modeled in terms of the cost of building that entity via combinations of other entities.   Suppose one has quantitative measures $\sigma: \mathcal{E} \rightarrow [0,\infty)$ and $\sigma^*: \mathcal{O} \rightarrow [0,\infty)$ (understood intuitively as measuring the simplicity of entities and combinatory operations respectively).   We will say that the pair $(\sigma, \sigma^*)$ is a CoSM if 

$$
\sigma(x) =  min_{y,z,i: x = y *_i z }  h(y,z)
$$

\noindent where

$$
h(y,z) = \sigma(y) + \sigma(z) + \sigma^*(*_i,y,z)
$$

\noindent where $\sigma^*(*_i,x,y) \equiv \sigma^*(\hat{*}_i)$ for the operation $y *_i z$.

Program length and program runtime are examples of COSMs.  Minimum program length to compute an entity $x$ in the programming language consisting of straightforward decision dags is a COSM that provides one measure of the number of distinctions one must make to compute the entity $x$.   Minimum program length in the CoDD programming language is a COSM that measures the number of distinctions one must make to compute $x$ leveraging reflection and substitution.   Worst-case runtime of the minimum-length decision dag or CoDD also yields a COSM.

This theory of CoSMs is extended to a theory of CoSMOS (combinatory Simplicity Measure Operating Sets) which involve multiple simplicity measures utilized together.  Given a vector of simplicity measures (aka a "multisimplicity measure"), an entity is associated not with an individual simplicity value but with a "simplicity bundles" of Pareto-optimal simplicity-value vectors.

CoDDs may be viewed as compositions of combinatory operators drawn from a vocabulary including conditionals, Boolean logic operators and a substitution operation.   The size of the most compact representation of a program as a CoDD is then precisely the simplicity of that program according the simplicity measure defined by these combinatory operations.  

A theory of pattern is then built up as follows: Let $(\vec{\sigma}, \vec{\sigma}^*) =(  (\sigma_1,\sigma_1^*), (\sigma_2,\sigma_2^*) ) $, where $(\sigma_1,\sigma_1^*)$ and $(\sigma_2,\sigma_2^*)$ are CoSMs with corresponding operator-sets $\mathcal{O}_1$, $\mathcal{O}_2$, with $\mathcal{O}_1 \subset \mathcal{O}_2$.   Denote $h_{1j}(y,z |w) = \sigma_1(y|w) + \sigma_1(z|w) + \sigma_j^{*|}(*_i,y,z|w)$ similarly to the definition of $h$ above (but noting that the first two terms use $\sigma_1$ and the third term $\sigma_j$).   Given this setup, we may define {\it pattern} as follows: the pair $(y,z)$ is a ${\bf pattern}$ in $x$ relative to multisimplicity measure $(\vec{\sigma}, \vec{\sigma}^*)$ and context $w$ with intensity (fuzzy degree)

$$
I_{y,z}^{(\vec{\sigma}, \vec{\sigma}^*)}(x|w) = \frac{\sigma_1(x|w) - h_{12}(y,z|w) } {\sigma_1(x|w)}
$$ 

\noindent We can then say that $(y,z)$ is a pattern in $x$ (relative to $w$) if the degree $I_{y,z}^{(\vec{\sigma}, \vec{\sigma}^*)}(x|w) > 0$. 

Next, the notion of a "subpattern hierarchy" is introduced, in which $x_i$ is a child of $x_k$ if there is some $x_j$ so that $x_i * x_j = x_k$ and $\sigma(x_i) + \sigma(x_j) < \sigma(x_k)$.  It is shown that if the combinatory operations by which the agents in the population underlying he computational model act on each other have a property called {\it mutual cost-associativity}, then the subpattern hierarchy has a transitivity property, i.e. if $x$ is a subpattern of $y$ and $y$ is a subpattern of $z$ then $x$ is a subpattern of $z$.   This provides an abstract understanding of how and why hierarchy is so often important in cognitive systems.   It is also pointed out that transitivity can be achieved by other means than associativity, e.g. if the agents are acting on a sufficient level of abstraction.  

\begin{figure}[htb]
\centering
\includegraphics[width=12cm]{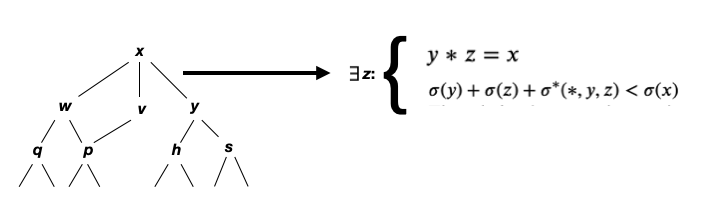}
\caption{Simple example of a subpattern hierarchy -- in which $y$ is a child of $x$ means there is some $z$ so that combining $y$ and $z$ together comprises a pattern in $x$.}
\label{fig:subpattern}
\end{figure}

\subsection{Associativity and Subpattern Hierarchy}

To formalize the notion of subpattern, we can define a binary operation $\leq$ on $\mathcal{E}$, the {\it subpattern relation} defined relative to $(\vec{\sigma}, \vec{\sigma}^*)$ and $w$ via

$$
x \leq y \iff max_z I_{x,z}^{(\vec{\sigma}, \vec{\sigma}^*)}(y|w) > 0
$$

\noindent If $x \leq y$, we will say that $x$ is a {\bf compositional subpattern} of $y$.  I.e., this means $x$ can be combined with some other entity $z$ to form a pattern in $y$.   

The notion of a {\it subpattern hierarchy} is then formally reflected by the assertion that, under reasonable conditions, the  subpattern relation is a {\it near partial order}, so that e.g. if $x \leq y$ and $y \leq z$ are both true then $x \leq z $ is almost true; and so that if $x \neq y$ and $x \leq y$ then it's not possible for $y \leq x$.  More formally,

\begin{mydef}
We will say that the subpattern relation $\leq$ is an {\bf approximate partial order} on ${\mathcal E}$ if it is reflexive and antisymmetric, and there is some constant $c>0$ so that

$$
x \leq y ,  y \leq z \rightarrow max_w I_{x,w}(z) \geq -c
$$

\end{mydef}

The simplest general-purpose way to obtain a subpattern hierarchy structure from a set of patterns is a property called {\it approximate cost-associativity}.   If the operator-set $*_i$ is mutually associative, then we will say that

\begin{mydef}
The mutually associative operator-set $\{*_i\}$ is {\bf approximately cost-associative}  relative to $\sigma$ if there is some constant $c>0$ so that

$$
| C_1(x,y,z) - C_2(x,y,z)    | < c
$$

\noindent where

\begin{itemize}
\item $C_1(x,y,z) = min_{i,j} ( \sigma^*(*_i, y, z) + \sigma^*(*_j, x, y *_i z))$ 
\item $C_2(x,y,z) = min_{i,j} ( \sigma^*(*_i, x*y, z) + \sigma^*(*_j, x, y)  )$
\end{itemize}

\end{mydef}

It is then shown in \cite{goertzel2020grounding} that:

\begin{thm} \label{thm:2}
The subpattern relation is an approximate partial order (with bound $c$) on $\mathcal{E}$ if:  The operations $*_i$ are approximately cost-associative (with bound $c$).  
\end{thm}

\subsubsection{From Subpattern Hierarchies to Dual Networks}

Extending the notion of a subpattern hierarchy further, in \cite{goertzel2020grounding} a formalization of the cognitive-systems notion of a "coherent dual network" interweaving hierarchy and heterarchy in a consistent way is presented.  A dual network, in this framework, is a network of agents where nodes that are nearby in the subpattern hierarchy have a high intensional similarity (are involved with a high percentage of overlapping patterns).   

Overall this direction of thinking re-envisions Occam's Razor as something like: When in doubt, prefer hypotheses whose simplicity bundles are Pareto optimal, partly because doing so both permits and benefits from the construction of coherent dual networks comprising coordinated and consistent multipattern hierarchies and heterarchies.

\subsection{Generalized Probabilities}

Perhaps the largest revolution in the AI field over the last few decades has been the rise of probabilistic methods.  The increasing amount of data readily available, via the Internet and improving low-cost sensors, has provided AI systems with sufficient data to carry out various sophisticated probabilistic inferences.   While some theorists have advocated focus on non-probabilistic methods of quantifying uncertainty (e.g. fuzzy methods \cite{Zadeh1978} or NARS \cite{Wang2006}), by and large probabilistic methods have carried the day due to their combination of demonstrated results and elegant mathematical footing.

Probabilistic methods commonly require prior distributional assumptions, which are then updated via observations.   The Solomonoff universal prior commonly used in algorithmic information theory \cite{Chaitin2008} may be viewed as a special case of a "simplicity prior", a probability distribution defined by normalizing a simplicity measure.   Simplicity thus leads naturally to probability as well as to pattern.

However, the standard approach to building probability distributions based on Boolean lattices is not the only relevant strategy from an AI point of view.  Knuth and Skilling's modern classic {\it Foundations of Inference} \cite{Knuth2012} paints a beautiful and vivid picture of probability as a quantitative representation of certain algebraic symmetries, and also makes clear that Boolean operations are not the only source of these pre-probabilistic symmetries.   Complex-valued quantum probabilities naturally ensue if one opts to represent uncertainties two dimensionally rather than one dimensionally; in \cite{goertzel2021paraconsistent} I have explored mappings between complex probability algebra and the 2D paraconsistent/ real-probablity algebra used in PLN.   And if one is interested in assigning probabilities to subgraphs of graphs or metagraphs, then one is naturally driven to looking at probability distributions defined on topologies of subgraphs.

The natural union, intersection and negation operations on subgraphs or submetagraphs form a Heyting algebra, and map isomorphically into the operations of intuitionistic logic \cite{goertzel2020paraconsistent}.    One may then naturally construct an intuitionistic probability theory based on the Heyting algebra of subgraphs.   Constructible Duality logic, a form of paraconsistent logic which as mentioned above is isomorphic to the PLN probabilistic logic used in OpenCog, is isomorphic to a pair of Heyting algebras.   By defining a probability theory on the open sets of this pair of Heyting algebras, one obtains an elegant grounding of PLN's uncertainty model.  

\begin{figure}[htb]
\centering
\includegraphics[width=12cm]{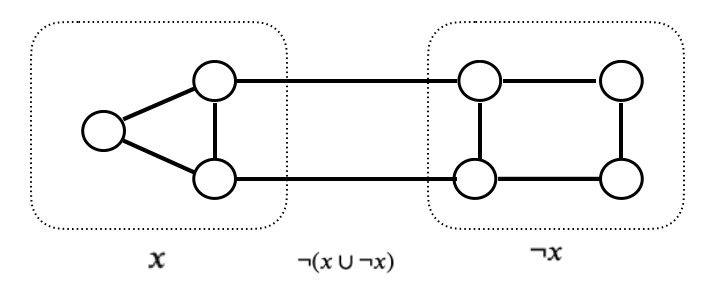}
\caption{Negation in the intuitionistic logic naturally associated with subgraphs.   The negation of a subgraph $G$ includes the nodes $N$ not in that subgraph and the links that interlink these among these nodes $N$, but not links between nodes of $G$ and $N$ -- which is why the Excluded Middle law doesn't apply.   Similar phenomena occur when deriving intensional logics based on sub-metagraphs.}
\label{fig:graph-neg}
\end{figure}

Godel's Second Incompleteness Theorem famously shows limitations to the ability of logical systems to reason consistently about themselves.   Paraconsistent and intuitionistic logics cannot entirely dodge this phenomenon.   However, it is possible for a logic system to carry out quite subtle and powerful reflective self-referential reasoning without falling into unproductive paradoxical situations in which the system totally loses ability to distinguish truth from falsehood, and appropriate use of paraconsistent and intuitionistic logic can help enable this.   One can map sets of equations in CD logic into non-well-founded sets (hypersets) as modeled by Aczel's Anti-Foundation Axiom (AFA) \cite{Aczel1988}; and correspondingly one can map sets of equations in weighted (uncertain) CD logic into infinite-order probability distributions defined over hypersets \cite{Goertzel2010q}, which as shown in \cite{goertzel2008mirror} can be used to construct interesting models of aspects of phenomenological experience such as self, will and reflective consciousness.

\begin{figure}[htb]
\centering
\includegraphics[width=12cm]{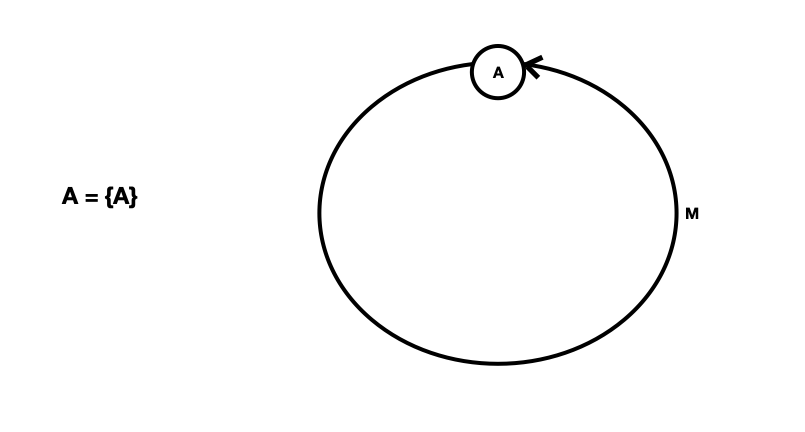}
\caption{Graphical depiction of the simplest "hyperset", aka anti-foundational set: A set that contains itself as its only element.  From \cite{Goertzel2011hyper}}
\label{fig:hyperset}
\end{figure}

\begin{figure}[htb]
\centering
\includegraphics[width=12cm]{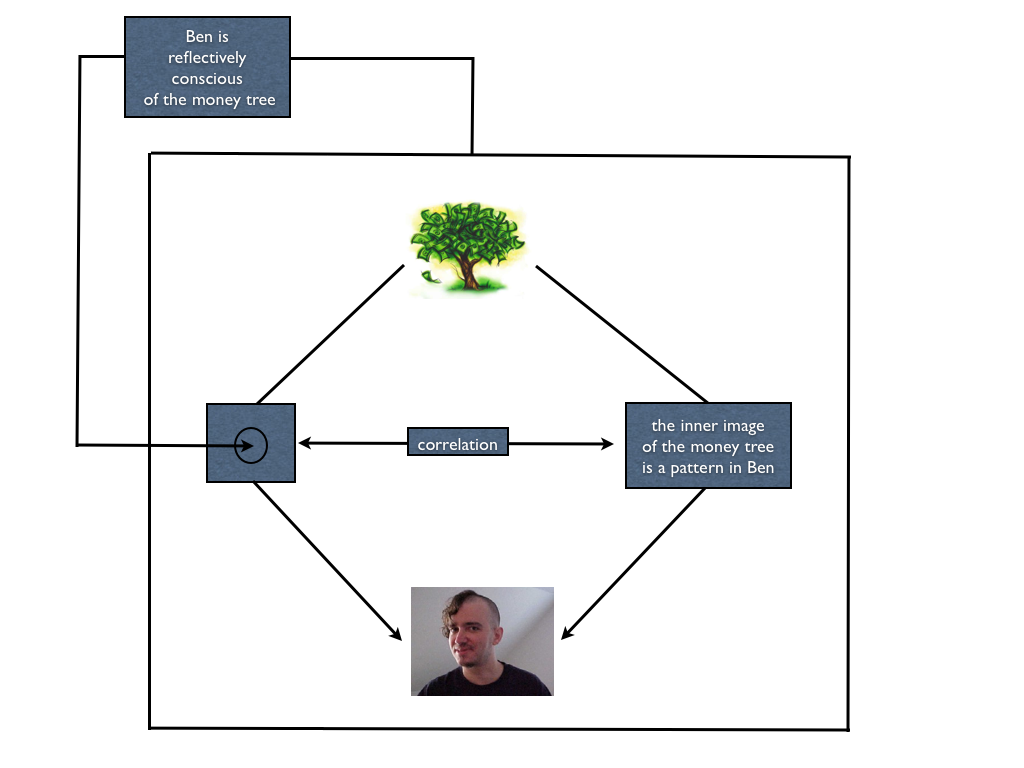}
\caption{Graphical depiction of a hyperset instantiating a simple logical model of reflective consciousness.  From \cite{Goertzel2011hyper}}
\label{fig:cons2}
\end{figure}

\begin{figure}[htb]
\centering
\includegraphics[width=12cm]{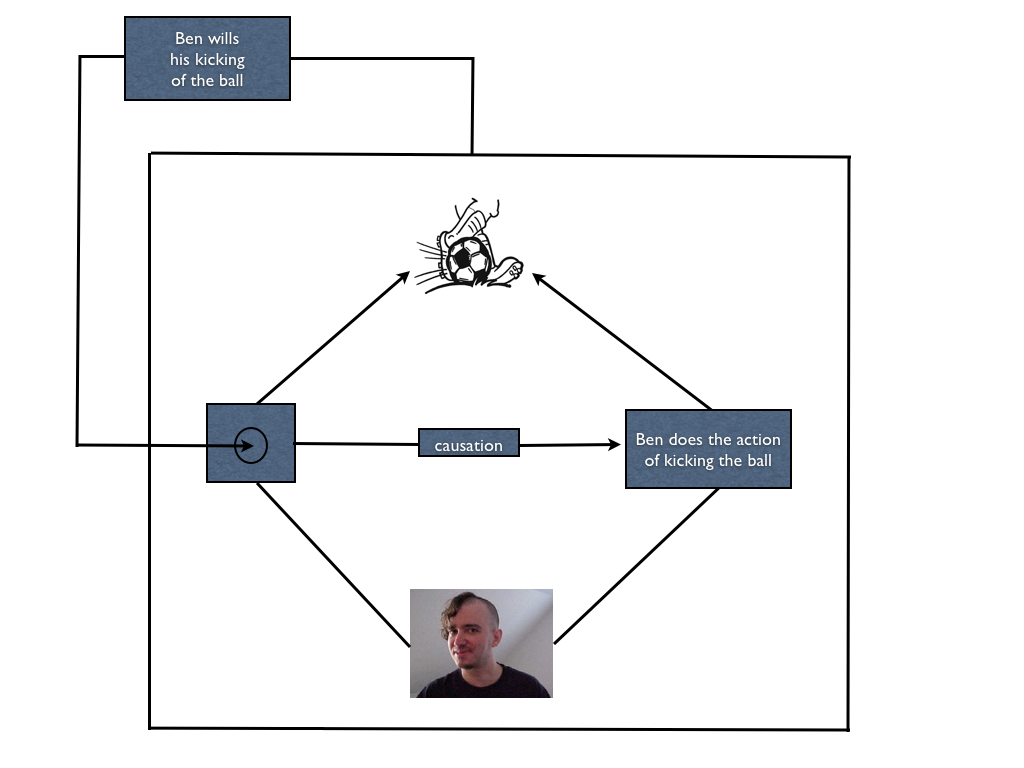}
\caption{Graphical depiction of a hyperset instantiating a simple logical model of the experience of willing.  From \cite{Goertzel2011hyper}}
\label{fig:cons6}
\end{figure}

\begin{figure}[htb]
\centering
\includegraphics[width=12cm]{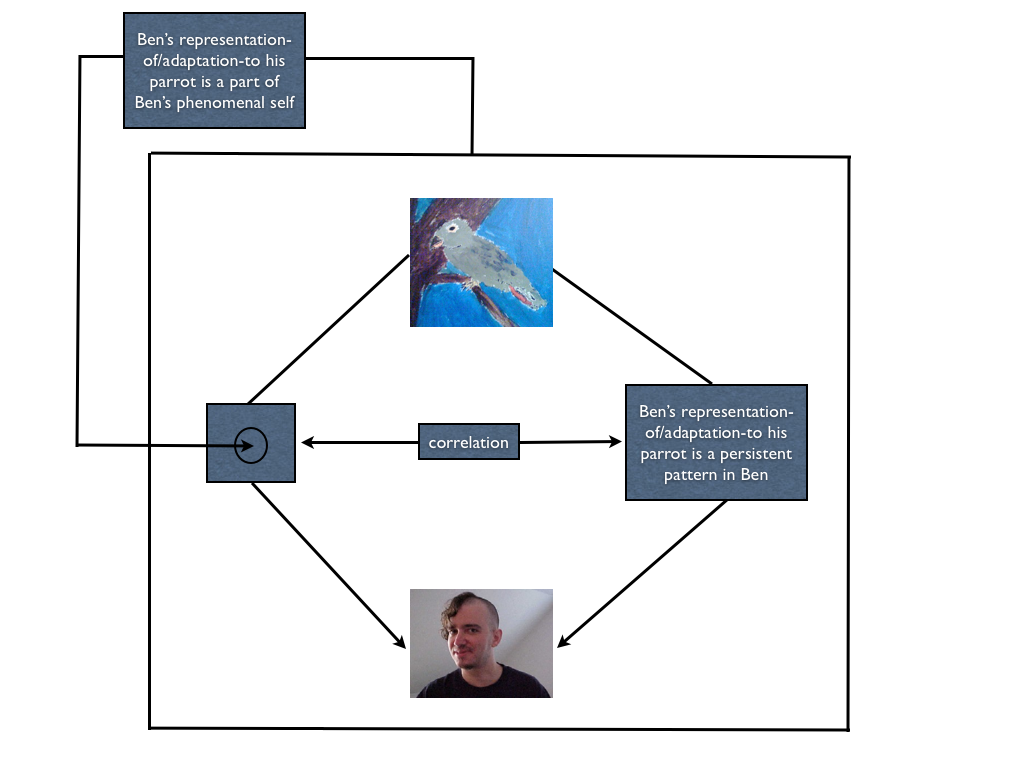}
\caption{Graphical depiction of a hyperset instantiating a simple logical model of the reflective self (the self which constructs itself as a model of itself). From \cite{Goertzel2011hyper}}
\label{fig:cons7}
\end{figure}

\section{Quantifying General Intelligence} \label{sec:def-intel}

Weaver's PhD thesis {\it Open-Ended Intelligence} \cite{weinbaum2017open} gives a beautiful and broad characterization of the nature of general intelligence, in essence viewing general intelligences as complex, self-organizing, self-constructing systems that recognize and form patterns in themselves and their environments.   

One can quantify the nature of generally intelligent systems in multiple ways.  For instance, "patternist ethics" identifies the three key values of Joy, Growth and Choice as applicable to multiple complex systems across multiple scales and contexts; these values may be quantitated relative to a specific definition of pattern and a specific local time-axis via refined versions of formulations such as

\begin{itemize}
 \item Joy is patterns persisting along the time-axis
 \item Growth is new pattern being created along the time axis
 \item Choice is a self-referential pattern of graphtropy decrease along the time axis
\end{itemize}

One can also quantitate various measures of "degree of intelligence" construed as e.g. general-purpose function optimization capability.  
Legg and Hutter \cite{Legg2007a} proposed a formal definition of intelligence, which we have extended in various ways in   \cite{goertzel2010toward}, and which is worthy of review and discussion in the present context.

\subsection{General Intelligence as Expected Reward Maximization Performance}

Following \cite{Legg2007}, we can make a simple formalization of the goal-achieving aspect of the intelligence by considering a class of active agents which observe and explore their environment and also take actions in it, which may affect the environment.  Formally, the agent sends information to the environment by sending symbols from some finite alphabet called the {\it action space} $\Sigma$; and the environment sends signals to the agent with symbols from an alphabet called the {\it perception space}, denoted $\mathcal P$.  Agents can also experience rewards, which lie in the {\it reward space}, denoted $\mathcal R$, which for each agent is a subset of the rational unit interval.  

The agent and environment are understood to take turns sending signals back and forth, yielding a history of actions, observations and rewards, which may be denoted $a_1 o_1 r_1 a_2 o_2 r_2 ...$ or else $a_1 x_1 a_2 x_2 ...$ if $x$ is introduced as a single symbol to denote both an observation and a reward.  The complete interaction history up to and including cycle $t$ is denoted $ax_{1:t}$; and the history before cycle t is denoted $ax_{<t}$ = $ax_{1:t-1}$.

The agent is represented as a function $\pi$ which takes the current history as input, and produces an action as output.  Agents need not be deterministic, an agent may for instance induce a probability distribution over the space of possible actions, conditioned on the current history.  In this case we may characterize the agent by a probability distribution $\pi( a_t  | ax_{<t} )$.    Similarly, the environment  may be characterized by a probability distribution $\mu(x_k | ax_{<k} a_k)$.  Taken together, the distributions $\pi$ and $\mu$ define a probability measure over the space of interaction sequences.

To define universal intelligence, Legg and Hutter consider the class of environments that are {\it reward-summable}, meaning that the total amount of reward they return to any agent is bounded by $1$.   Where $r_i$ denotes the reward experienced by the agent from the environment at time $i$, the {\it expected total reward} for the agent $\pi$ from the environment $\mu$ is defined as

$$
V_{\mu}^{\pi} \equiv {E  ( \sum_1^{\infty} r_i} ) \leq 1
$$

\noindent where $K(\mu)$ is the Kolmogorov complexity (which denotes, essentially, the length of the shortest program computing $\mu$, Legg and Hutter define

\begin{mydef} [{\bf Legg and Hutter}]
The {\bf universal intelligence} of an agent $\pi$ is its expected performance with respect to the universal distribution $2^{-K(\mu)}$ over the space of all computable reward-summable environments, $E$, that is, as

$$
\Upsilon(\pi) \equiv \sum_{\mu \in E} ( 2^{-K(\mu)} V_{\mu}^{\pi} )
$$
\end{mydef}

\noindent and they point out that $\Upsilon (\pi) = V_{\xi}^{\pi}$ where $\xi$ is the universal distribution implied by the Kolmogorov complexity, which means that, as they phrase it, "the universal intelligence of an agent is simply its expected performance with respect to the universal distribution."

\subsection{Pragmatic General Intelligence}

In \cite{goertzel2010toward} I consider a slightly generalized version of Legg and Hutter's definition of general intelligence called "Pragmatic General Intelligence," broken down to consider goals and environments separately and to encompass more general priors than the Solomonoff prior.  I introduce the notion of a {\it goal}, meaning a function that maps finite sequences $ax{s:t}$ into rewards.  As well as a distribution over environments, we have need for a conditional distribution $\gamma$, so that $\gamma(g,\mu)$ gives the weight of a goal $g$ in the context of a particular environment $\mu$. 

A {\it goal-seeking agent} is considered as agent that receives an additional kind of input besides the perceptions and rewards considered above: it receives goals.  In this extended framework, an interaction sequence looks like $m_1 a_1 o_1 g_1 r_1 m_2 a_2 o_2 g_2 r_2  ...$ or else $w_1 y_1 w_2 y_2 ...$ if $w$ is introduced as a single symbol to denote the combination of a memory action and an external action, and $y$ is introduced as a single symbol to denote the combination of an observation, a reward and a goal.   It is assumed that the reward $r_i$ provided to an agent at time $i$ is determined by the goal function $g_{i}$. 

A goal may come with a natural time-scale, which is represented as a Boolean indicator function over the integers.  The Boolean value $\tau_{g,\mu}(n)$ tells whether it makes sense to evaluate performance on goal $g$ in environment $\mu$ over a period of $n$ time steps ($1$ means yes, $0$ means no).  The term "context" is used here to denote the combination of an environment, a goal function and a reward function.

If the agent is acting in environment $\mu$, and is provided with $g_t = g$ for the time-interval $T = t \in \{t_1 ,..., t_2\}$, then the {\it expected goal-achievement} of the agent during the interval is the expectation

$$
V_{\mu, g,T}^{\pi} \equiv  E(  \sum_{t_1}^{t_2} r_i )
$$

One may introduce a second-order probability distribution $\nu$, which is a probability distribution over the space of environments $\mu$.   One may then say

\begin{mydef}
The {\bf pragmatic general intelligence} of an agent $\pi$, relative to the distribution $\nu$ over environments and the distribution $\gamma$ over goals, is its expected performance with respect to goals drawn from $\gamma$ in environments drawn from $\nu$, over the time-scales natural to the goals; that is,

$$
\Pi(\pi) \equiv \sum_{\mu \in E, g \in {\mathcal G}, T}  \nu(\mu) \gamma(g, \mu) \tau_{g,\mu}(|T|) V_{\mu,g,T}^{\pi} 
$$

\noindent where $|T|$ denotes the length of the time-interval $T$ (and in those cases where this sum is convergent).
\end{mydef}

This definition formally captures the notion that "intelligence is achieving complex goals in complex environments," where "complexity" is gauged by the assumed measures $\nu$ and $\gamma$.

A further step is to incorporate an agent's resource usage into the picture.   Let $\eta_{\mu,g,T}$ be a probability distribution describing the amount of computational resources consumed by an agent while achieving goal $g$ over time-scale $T$.  This is a probability distribution because we want to account for the possibility of nondeterministic agents.   So, $\eta_{\mu,g,T}(Q)$ tells the probability that $Q$ units of resources are consumed.  For simplicity we amalgamate space and time resources, energetic resources, etc. into a single number $Q$, which is assumed to live in some subset of the positive reals.   Space resources of course have to do with the size of the system's memory, briefly discussed above.  Then we may define

\begin{mydef}
The {\bf efficient pragmatic general intelligence} of an agent $\pi$ with resource consumption $\eta_{\mu,g,T}$, relative to the distribution $\nu$ over environments and the distribution $\gamma$ over goals, is its expected performance with respect to goals drawn from $\gamma$ in environments drawn from $\nu$, over the time-scales natural to the goals, normalized by the amount of computational effort expended to achieve each goal; that is,

$$
\Pi_{Eff}(\pi) \equiv \sum_{\mu \in E, g \in {\mathcal G}, T, Q}  \frac {\nu(\mu) \gamma(g, \mu) \tau_{g,\mu}(|T|) \eta_{\mu,g,T} (Q) } { Q} V_{\mu,g,T}^{\pi} 
$$

(in those cases where this sum is convergent).
\end{mydef}

\noindent Efficient pragmatic general intelligence is a measure that rates an agent's intelligence higher if it uses fewer computational resources to do its business.

Another approach to incorporating computational resource usage into the quantification of general intelligence would be to shift to a multiobjective optimization framework and consider minimization of time, space and energetic resource utilization as objective functions to be balanced alongside expected degree of achievement of other goals, e.g. in a Pareto-optimization based framework.

\subsection{Intellectual Breadth}

One can also define the {\it generality} or {\it breadth} of an intelligent system's function optimization capability, which is largely orthogonal to its degree of optimization capability.   To formalize this simply, consider "contexts" that are constructed as "environment/interval triple $(\mu, g,T)$."   Given a context $(\mu,g,T)$, and a set $\Sigma$ of agents, one may construct a fuzzy set $Ag_{\mu,g,T}$ gathering those agents that are intelligent relative to the context; and given a set of contexts, one may also also define a fuzzy set $Con_{\pi}$ gathering those contexts with respect to which a given agent $\pi$ is intelligent.  The relevant formulas are:

$$
\chi_{Ag_{\mu,g,T}}(\pi) = \chi_{Con_{\pi}}( \mu,g,T ) = \sum_{Q} \frac{ \eta_{\mu,g,T} (Q) V_{\mu,g,T}^{\pi}}{Q} 
$$

\noindent  One can then say

\begin{mydef}
The {\bf intellectual breadth} of an agent $\pi$, relative to  the distribution $\nu$ over environments and the distribution $\gamma$ over goals, is

$$
H( \chi^P_{Con_{\pi}}( \mu,g,T) )
$$

\noindent where $H$ is the entropy and

$$
\chi^P_{Con_{\pi}}( \mu,g,T ) = 
$$

$$
 \frac { \nu(\mu) \gamma(g, \mu) \tau_{g,\mu}(|T|) \chi_{Con_{\pi}}( \mu,g,T )}{\sum_{(\mu_{\alpha},g_{\beta}.T_{\omega}) } \nu(\mu_{\alpha}) \gamma(g_{\beta}, \mu_{\alpha}) \tau_g(|T_{\omega}|) \chi_{Con_{\pi}}( \mu_{\alpha},g_{\beta},T_{\omega} ) } 
$$

\noindent is the probability distribution formed by normalizing the fuzzy set $\chi_{Con_{\pi}}( (\mu,g.T) )$.
\end{mydef}

\subsection{Multiple Criterion Driven General Intelligence}

The relationships between joy, growth, choice, breadth, goal-achievement and efficient resource utilization in complex systems are subtle and currently not very well understood.   However it seems clear that real-world general intelligences should not be understood or engineered as simple single-utility-function maximizers.   At a rough initial approximation, it seems we should think in terms of configuring our early-stage AGI systems to concurrently pursue multiple objectives including versions of joy, growth and choice, as well as more concrete goals such as survival and safety for humans.    Given the leeway any proto-AGI system will inevitably have in interpreting such goals and grounding them in real-world situations, and the flexibility an advanced AGI system will need to have in revising and improving its own code including its goal system, it's clear that the formalization of objectives can be meaningfully considered only alongside the practical situations in which the AGI systems will be embedded as it grows.

\section{Universal Algorithms for General Intelligence}

Marcus Hutter's classic work {\it Universal AI} \cite{Hutter2005} presents a "universal AGI process" called AIXI, which in a sense provides a thorough and optimal solution to the problem of maximizing an arbitrary computable reward function in an arbitrary computable environment.  AIXI is itself uncomputable, but has computable approximations such as AIXI$^{tl}$ that are computable-in-principle but merely completely unrealistic to compute.   Very roughly speaking the way AIXI$^{tl}$ works is: At each step it brute-force searches the space of all computer programs of length $\leq l$ and runtime $\leq t$ and finds the shortest program $P$, among these, that maximizes the expected reward conditional on execution of $P$ to generate the agent's next step.   The prediction of expected reward is done by probabilistic reasoning with a prior distribution that assigns greater prior probability to programs with shorter length.   

AIXI$^{tl}$  is completely infeasible to implement in practice, but it gives a way of thinking about practical AGI algorithms.   A practical AGI algorithm can be viewed as doing something similar but replacing the brute-force search with heuristic search that is, on average, especially effective in the context of the particular reward functions and environments that a certain agent is especially concerned with.
One is then led down the path of exploring and formalizing the properties of the goals and environments actually encountered by real intelligent agents achieving real goals in real physical, social and intellectual worlds, and how these map into properties of heuristic search algorithms.

Schmidhuber's Godel Machine \cite{Schmidhuber2006} provides a different twist on the same idea.   Roughly speaking: One looks at an AGI system supplied with a certain formal logic, and then asks the system to choose its next action $A$ by using its logic and its available data to prove that $A$ is the action that will provide it the maximum expected reward.   This approach can be applied to internal actions as well as external actions, making it a recursive approach to probabilistic inference control.   Searching over proofs of arbitrary length gives a system conceptually similar to AIXI, whereas searching over proofs of bounded length gives a system conceptually similar to AIXI$^{tl}$.   While such a parallel hasn't been elaborated formally so far as I know, it seems that AIXI$^{tl}$ and the bounded-proof-length Godel Machine must be tied together via a Curry-Howard type correspondence, of the same sort that is used to establish isomorphism between program-execution and proof in so many other contexts.

The basic concept of these universal AI approaches can be generalized to a framework where one has multiple goal functions, which need not be expressible as expected reward functions, but can simply be mappings from future histories to real numbers.    Given a set of such goal functions and a set of simplicity measures, one can look an hypothetical AGI system that brute-force searches the space of all computer programs of length $\leq l$ and runtime $\leq t$ to find those that are Pareto-optimal for the simplicity measures, among those that are Pareto-optimal for the goal functions.   Or one can look at a logic-based AGI system that strives to take actions that are provably (with proofs below some fixed length) Pareto-optimal for the simplicity measures, among those that are Pareto-optimal for the goal functions.  Computational resource restrictions can be baked into the goal framework as hinted above, with minimization of space, time or energetic complexity as goals in the mix.

\subsection{General World-Modeling Principles for General Intelligence}

It is interesting to ask how -- or in what sense -- these hypothetical arbitrarily-powerful AGI systems model the world as they go about making their decisions of what actions to take.   Of course the brute-force search algorithms involved in methods such as AIXI$^{tl}$  and the Godel Machine don't do any explicit world-modeling -- but their actions may nevertheless be implicitly consistent with certain sorts of world-models, and looking at what these are can be useful in crafting realistic approximations of these abstract algorithms.

It appears that rearranging the arithmetic of evidence counting in an appropriate way allows one to formulate general-purpose world-modeling principles that, in a certain sense, every sufficiently powerful intelligent system will do well to at least roughly approximate in its quest to understand itself and the world.

As the first step down this path, consider that: The Maximum Entropy Principle (MaxEnt) allows one to infer the most likely probability distribution for the variables characterizing a system given a set of linear constraints on that state -- via choosing the distribution that has the highest entropy among those consistent with the constraints \cite{Jaynes2003}.   Basically this is the distribution whose description requires the least amount of additional statistical information beyond the information in the constraints themselves.

The Maximum Caliber Principle (MaxCal) \cite{dixit2018perspective} extends MaxEnt to systems that change over time -- basically it says that given linear constraints on a system that probabilistically evolves over time, the most likely probability distribution over system trajectories is the one that maximizes the entropy in trajectory-set space (the "caliber").   Just as MaxEnt can be generalized to graphtropy rather than entropy, so can MaxCal, via creating distinction graphs embedding distinctions between trajectories.

\begin{figure}[htb]
\centering
\includegraphics[width=12cm]{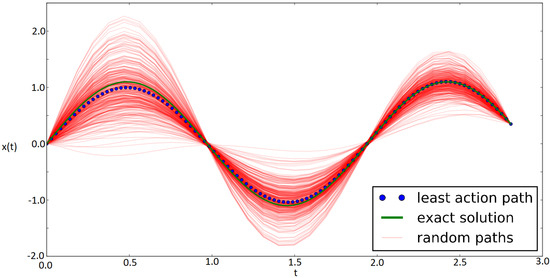}
\caption{Physics example of the Maximum Caliber Principle, used to guide Monte Carlo sampling to find the most probable path of a harmonic oscillator with fixed kinetic foci.  From \url{https://www.mdpi.com/1099-4300/22/9/916/htm}}
\label{fig:maxcal}
\end{figure}

The relevance of these principles to AGI is: These are deeply mathematically grounded heuristics that any intelligent system will do well to use when grappling with its complex, uncertain world.

The analogue of MaxEnt in the realm of algorithmic rather than statistical information involves Algorithmic Markov processes \cite{janzing2010causal}, the algorithmic-information analogue of ordinary statistical Markov processes.   The action of an Algorithmic Markov process turns out to be the most rational hypothesis to use when inferring underlying structures based on data.   Intuitively, if you looked at the patterns in the choices of an AIXI$^{tl}$ type agent  over time, you would see that the system was implicitly making the assumption that the world is often roughly built via an an algorithmic Markov process, conditional on its knowledge about the world.   Assuming algorithmic Markovicity depending on observed constraints, on the part of a process constructing an observed entity, is basically equivalent to assuming independence between constructive processes that are not specifically known to be dependent, because there are more ways for the processes to be independent than there are ways for them to be dependent {\it in any particular way} (and by assumption one doesn't have knowledge about any particular dependency between the processes).

I have argued in \cite{goertzel2019maximal} that MaxCal can similarly be extended to a "maximum algorithmic caliber principle" that characterizes the possible worlds most likely to accord with a given set of observations -- one should assume the world has evolved with the maximum algorithmic caliber consistent with observations (basically, the most computationally dense way consistent with observations).   Basically, this just means that in hypothesizing the processes underlying some temporal observations, you should assume independence between subprocesses that are not specifically known to be dependent, because there are more ways for the processes to be independent than there are ways for them to be dependent {\it in any particular way}.   

One interesting point here is that assuming a simplicity prior leads to inference principles that involve assigning maximal likelihood to the possible worlds that are in a sense maximally complex.   However there is no contradiction here, just a subtlety.   The simplicity prior is about how the conditional "information" (the conditional simplicity or complexity) of one entity or process is calculated relative to another -- one calculates this by looking at the simplest way to get from the one to the other, using the assumed COSM (e.g. the assumed underlying programming language such as CoDD).   Given this model of inter-transformations between entities and processes, one can then look at the scope of models of the world, and one finds that the greatest volume of models consistent with observation exists in the vicinity of the Algorithmic Markov dag constructible from observations based on the given simplicity measure.   

Like traditional MaxEnt and MaxCal, these algorithmic versions are also deeply mathematically grounded heuristics that any sufficiently intelligent system will do well to use -- explicitly or implicitly -- when grappling with its complex, uncertain world.   However they are also profoundly computationally intractable in their pure form.  So it's more accurate to say that any intelligent system will do well to explicitly or implicitly {\it roughly approximate} the use of these heuristics, using approximations that appropriately match its own environment and nature.

Why is this interesting?   Along with the pure intellectual interest, because: When thinking about practical approximations to intractable idealistic AGI algorithms, it is sometimes interesting to also think in terms of practical approximations to intractable idealistic world-models like algorithmic Markov processes.

\section{Specializing Maximally General AGI via Combining Practical Discrete Decision Systems} \label{sec:metagraph-agi}

AIXI, Godel Machine and their relatives have the general form of "reinforcement learning" or "experiential interactive learning" algorithms, meaning that they operate via iteratively observing the world (including observing the impact of their own actions on the world), then choosing actions that they expect will give them maximum reward based on the world's reactions, etc.   They are unrealistic because their methodology of action selection is uncomputable or else (in the simplistic approximate versions of the original uncomputable algorithms)  computationally intractable.  This leaves open the question whether there are meaningful ways to "scale down" from these intractable algorithms toward more feasible algorithms that somehow preserve the spirit of the original fully general but uncomputable or intractable approaches.   Hutter and Schmidhuber and colleagues have pursued a variety of research explicitly aimed in this direction, e.g. Monte Carlo AIXI \cite{Hutter:11aixictwx} and OOPS (Optimal Ordered Problem Solver) \cite{schmidhuber2004optimal} and Arthur Franz's work mentioned above as examples; however these efforts so far have not thoroughly or richly connected with the world of practical AI algorithms and systems.

In a recent paper {\it Patterns of Cognition} \cite{goertzel2021patterns} I have sought to provide one sort of conceptual and mathematical bridge between these infeasible general-purpose AGI frameworks and practical real-world AGI-oriented systems, via looking at formulations of the AI algorithms playing key roles in the OpenCog AI system in terms of abstract recursive discrete decision systems.   These DDSs (Discrete Decision Systems) on the one hand can be straightforwardly viewed as scaled down versions of AIXI$^{tl}$ or time-bounded Godel Machine type systems, but on the other hand can be used to drive concrete thinking about functional programming implementations of OpenCog algorithms.        

The {\it Patterns of Cognition} analysis involves representing various cognitive algorithms as recursive discrete decision processes involving optimizing functions defined over metagraphs, in which the key decisions involve sampling from probability distributions over metagraphs and enacting sets of combinatory operations on selected sub-metagraphs.    A variety of recursive decision process called a COFO (Combinatory Function Optimization) algorithm plays a key role.   One can view a COFO as being vaguely like Monte-Carlo-AIXI, but within the context of a combinatory computational model -- and with the added twist that the Monte Carlo sampling based estimations are augmented by estimations using other probabilistic algorithms that are themselves implemented using COFO.   There are close connections to modern probabilistic programming theory \cite{wood2018-probprog}, but with more of an emphasis on recursive inference algorithms and less reliance on simplistic sampling methods.

Behind the scenes of the COFO framework is a core insight drawn from the body of theory behind the OpenCog system -- that a combinatory computational model defined over metagraphs is an especially natural setting in which to formalize various practical AGI-oriented algorithms.   From a sufficiently abstract perspective, all Turing-complete computational models are equivalent, and all general-purpose computational data structures are equivalent.  But from a practical AGI implementation and teaching perspective, it makes a difference which computational models and data structures one chooses; the argument for metagraphs as the core data structure for AGI has been laid out in \cite{Goertzel2020metagraph} and references therein such as \cite{gibbons1995initial}, and the argument for combinatory computing as the core approach for AGI has been laid out in \cite{EGI1} and earlier papers referenced therein.

\subsection{Discrete Decision Systems}

\begin{figure}[htb]
\centering
\includegraphics[width=12cm]{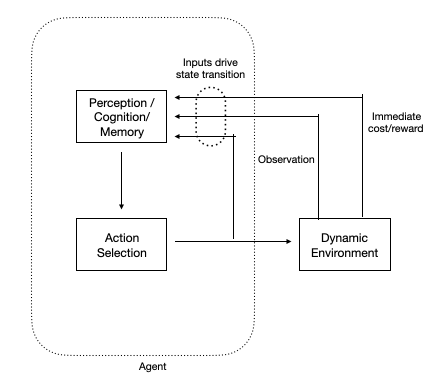}
\caption{Schematic diagram of a fairly general Discrete Decision System for controlling a computational agent.   This model is general enough to cover AIXI$^{tl}$ type systems as well as classic dynamic programming systems, Actor-Critic style reinforcement learning systems, and broader experience-driven cognitive architectures like OpenCog.}
\label{fig:dds}
\end{figure}

To bridge the gap between abstract agent models like the ones considered in Section \ref{sec:def-intel} above and practical AGI systems, we introduce a basic model of a {\it discrete decision system (DDS)} -- a process depicted in Figure \ref{fig:dds} and formally defined on $n$ stages in which each stage $t=1,\ldots,n$ is characterized by

\begin{itemize}
\item an {\bf initial state} $s_t\in S_t$, where $S_t$ is the set of feasible states at the beginning of stage $t$;
\item an {\bf action} or '''decision variable''' $x_t\in X_t$, where $X_t$ is the set of feasible actions at stage $t$ -- note that $X_t$ may be a function of the initial state $s_t$;
\item an {\bf immediate cost/reward function} $p_t(s_t,x_t)$, representing the cost/reward at stage $t$ if $s_t$ is the initial state and $x_t$ the action selected;
\item a  {\bf state transition function}  $g_t(s_t,x_t)$ that leads the system towards state $s_{t+1}=g_t(s_t,x_t)$.
\end{itemize}

\noindent  The mapping of the simple agents model given above into this framework is fairly direct: environments determine which actions are feasible at each point in time and goals are assumed decomposable into stepwise reward functions.  Highly generally intelligent agents like AIXI$^{tl}$ fit into this framework, but so do practical AI algorithm frameworks like greedy optimization and deterministic and stochastic dynamic programming.   As we shall see, with some care and further machinery the various cognitive algorithms utilized in the OpenCog framework can be interpreted as DDSs as well.

To express greedy optimization in this framework, one begins with an initial state, chosen based on prior knowledge or via purely randomly or via appropriately biased stochastic selection.   Then one chooses an action with a probability proportional to immediate cost/reward (or based on some scaled version of this probability).  Then one enacts the action, the state transition, and etc.  

An interesting case of "greedy" style DDS dynamics in an AGI context is the adaptive spreading of attention through a complex network.   OpenCog's attentional dynamics subsystem, ECAN (Economic Attention Networks), involves spreading of two types of attention values through a knowledge metagraph -- Short-Term Importance (STI) and Long-Term Importance (LTI) values, representing very roughly the amount of processor time an Atom should receive in the near term, and the critical-ness of keeping an Atom in RAM in the near term.   In this case

\begin{itemize}
\item an {\bf initial state} is a distribution of STI and LTI values across the Atoms in an Atomspace
\item an {\bf action} is the spreading of some STI or LTI from one Atom to its neighbors
\item an {\bf immediate cost/reward function} is the degree to which a given spreading action causes the distribution of STI/LTI values to better approximate the actual expected utilities of assignation of processor time and RAM to the Atoms in Atomspace
\item a  {\bf state transition function} is the updating of the overall set of STI/LTI values
\end{itemize}

\noindent and the ECAN equations in the OpenCog system embody a greedy heuristic for executing this DDS.

To express dynamic programming in this DDS framework is a little subtler, as in DP one tries to choose actions with probability proportional to overall expected cost/reward.  Estimating the overall expected cost/reward of an action sequence requires either an exhaustive exploration of possibilities (i.e. full-on dynamic programming) or else some sort of heuristic sampling of possibilities (approximate stochastic dynamic programming).   One can also take a curve-fitting approach and try to train a different sort of model to approximate what full-on dynamic programming would yield -- e.g. a model that suggests series of actions based on a holistic analysis of the situation being confronted, rather than proceeding explicitly through an iterative sequence of individual actions and decisions.

AIXI$^{tl}$ can be viewed as a way of doing approximate stochastic dynamic programming that minimizes the degree of error given assumption of a Solomonoff prior and given the assumption of access to an oracle that rapidly searches the space of all programs of suitably bounded length and runtime.   Generalizing to cope with different sorts of priors is not so complex if one is willing to stick with the assumption of availability of massive computational resources, and focus on the long-term intelligence of an agent that has made and adapted to a large number of observations.   Generalizing to deal with realistic resource restrictions and the unavailability of near-magical oracles is a bigger deal and drives one toward algorithms that leverage specific assumptions about goals and environments (baked in the DDS framework into specific reward functions) via specific memory and algorithmic structures.

A quick note about parallel and distributed processing before going further: While it's simplest to think about this sort of DDS in terms of one action being executed at a time, the framework as stated is general enough to encompass concurrency as well.   One can posit underlying atomic actions $w_t \in W_t$, and then define the members of $X_t$ as subsets of $W_t$.   In this case each action $x_t$ represents a set of $w_t$ being executed concurrently.

\subsection{Combinatory-Operation-Based Function Optimization} \label{sec:COFO}

To frame the sorts of cognitive algorithms involved in OpenCog and related AGI architectures in terms of general DDS processes, \cite{goertzel2021patterns} introduces the notion of COFO, Combinatory-Operation-Based Function Optimization.   Basically, a COFO process, as depicted in Figure \ref{fig:cofo} wraps a combinatory computational system of the sort considered in Section \ref{sec:pattern} above within a DDS, by using the combinatory system as the method of choosing actions in a discrete decision process oriented toward optimizing a function.   The hypothesis is then made that this particular sort of DDS plays a core role in practical AGI systems operating in environments relevant to our physical universe and the everyday human world.

More specifically, we envision a cognitive system controlling an agent in an environment to be roughly describable as a DDS (the "top-level DDS"), and then envision the cognitive processing in the Perception / Inference / Memory box in the DDS diagram in Figure \ref{fig:dds} as comprising:

\begin{itemize}
\item A memory consisting of a set of entities that combine with each other to produce other entities, i.e. a combinatory system embodied in a knowledge metagraph
\item Cognitive processes instantiated as COFO processes, i.e as DDSs whose goals are function optimizations and whose actions are function evaluations, all leveraging a common metagraph as background knowledge and as a dynamic store for intermediate state
\item One or more DDSs carrying out attention allocation on the common metagraph (the core DDS here using greedy heuristics but supplemented by one or more additional DDSs using more advanced cognition), spanning the portions of the metagraph focusedon by the various COFO processes
\end{itemize}

\noindent So practical intelligent systems are modeled as multi-level DDSs where the subordinate DDSs operating within the outer-loop agent control DDS are mostly COFO processes.   Section \ref{sec:cogarch} below explores how the various COFO-like processes involved in human-like cognition appear to interoperate in human cognitive architecture, and Section \ref{sec:humancog} explores more specifically how the OpenCog Hyperon design explicitly interleaves COFO processes in its attempt to manifest advanced AGI.

A COFO process, more explicitly, involves making of a series of decisions involving how to best use a set of combinatory operators $C_i$ to gain information about maximizing a function $F$ (or Pareto optimizing a set of functions $\{F_i\}$) via sampling evaluations of $F$ ($\{F_i\}$).   For simplicity we'll present this process in the case of a single function $F$ but the same constructs work for the multiobjective case.   It is shown in \cite{goertzel2021patterns} how COFO can be represented as a discrete decision process, which can then be enacted in greedy or dynamic programming style.

\begin{figure}[htb]
\centering
\includegraphics[width=12cm]{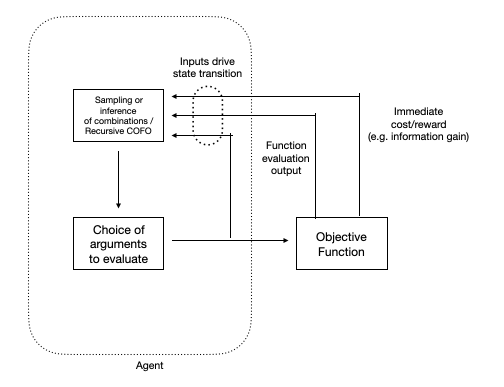}
\caption{Schematic diagram of the Combinatory-Operation-Based Function Optimization (COFO) process for optimizing functions via searching spaces of combinations.   COFO processes follow the general template of DDS but the actions they are concerned with are actions of providing an argument to be evaluated by a partially-unknown function that is the subject of optimization activity.   The core algorithms involved in AGI systems like OpenCog can be expressed as cases of the COFO process.}
\label{fig:cofo}
\end{figure}

\begin{figure}[htb]
\centering
\includegraphics[width=12cm]{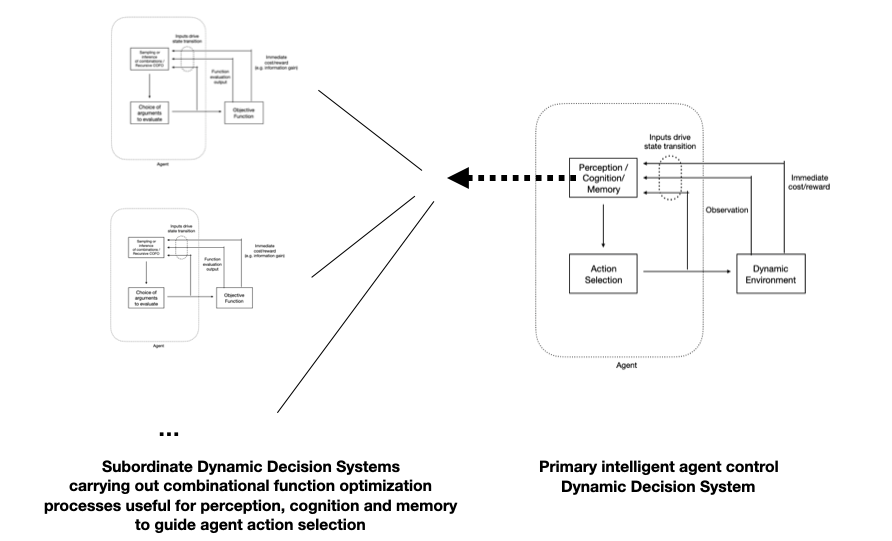}
\caption{Schematic diagram illustrating how DDSs implementing COFO processes may be leveraged within a top-level DDS used to control an intelligent agent acting in an environment.   The COFO DDSs are engaged to support the top-level DDS in cognition, perception, memory and action selection -- leveraged for their particular function optimization capabilities according to the agent's overall cognitive architecture.}
\label{fig:cofo}
\end{figure}

Given a function $F: X \rightarrow R$ (where $X$ is any space with a probability measure on it and $R$ is the reals), let $\mathcal{D}$ denote a "dataset" comprising finite subset of the graph $\mathcal{G}(F)$ of $F$, i.e. a set of pairs $(x,F(x))$.   We want to introduce a measure $q_F(\mathcal{D})$ which measures how much guidance $\mathcal{D}$ gives toward the goal of finding $x$ that make $F(x)$ large.   The best measure will often be application-specific; however as shown in \cite{goertzel2021patterns} one can also introduce general-purpose entropy-based measures that apply across domains and problems.

We can then look at greedy or dynamic programming processes aimed at gradually building a set $D$ in a way that will maximize  $q_{\rho,F}(D)$.   Specifically, in a cognitive algorithmics context it is interesting to look at processes involving combinatory operations $C_i: X \times X \rightarrow X$ with the property that 

$$
P( C_i(x, y  ) \in M_\rho^D | x \in M_\rho^D, y \in M_\rho^D)  \gg P(z \in M_\rho^D | z \in X)
$$

\noindent That is, given $x, y \in M_\rho^D$, combining $x$ and $y$ using $C_i$ has surprisingly high probability of yielding $ z \in M_\rho^D$.

Given combinatory operators of this nature, one can then approach gradually building a set $D$ in a way that will maximize  $q_{\rho,F}(D)$, via a route of successively applying combinatory operators $C_i$ to the members of a set $D_j$ to obtain a set $D_{j+1}$.

Framing this COFO process as a form of recursive Discrete Decision System (DDS), we obtain:

\begin{enumerate}
\item A {\bf state} $s_t$ is a dataset $D$ formed from function $F$
\item  An {\bf action} is the formation of a new entity $z$ by 
\begin{enumerate}
\item Sampling $x, y$ from $X$ and $C_i$ from the set of available combinatory operators,  in a manner that is estimated likely to yield $z=C_i(x,y)$  with $z \in M_\rho^D$
\begin{enumerate}
\item As a complement or alternative to directly sampling, one can perform probabilistic inference of various sorts to find promising $(x,y,C_i)$.   This probabilistic inference process itself may be represented as a COFO process, as we show below via expressing PLN forward and backward chaining in terms of COFO
\end{enumerate}
\item Evaluating $F(z)$, and setting $D^* = D \cup (z, F(z))$.
\end{enumerate}
\item  The {\bf immediate reward} is an appropriate measure of the amount of new information about making $F$ big that was gained by the evaluation $F(z)$.   The right measure may depend on the specific COFO application; one fairly generic choice would be the relative entropy $q_{\rho,F}(D^*, D)$ 
\item  {\bf State transition}: setting the new state $s_{t+1}=D^*$
\end{enumerate}

\noindent A concurrent-processing version of this would replace 2a with a similar step in which multiple pairs $(x,y)$ are concurrently chosen and then evaluated.

The action step in a COFO process is in essence carrying out a form of probabilistic programming \cite{wood2018-probprog} (which is clear from the discussion of probabilistic programming in a dependent type context given in \cite{goertzel2020paraconsistent}).   Finding the right conglomeration of combinatory operators to produce a given output is formally equivalent to finding the right program to produce a given sort of output, and here as in probabilistic programming one is pushed to judiciously condition estimates on prior knowledge.  There are big practical differences from most current probabilistic programming systems -- the simplified distributional assumptions commonly made are not wired in here, and sophisticated recursive inference is explicitly relied upon, rather than mostly defaulting to simple non-scalable sampling methods.  But on a conceptual and formal level COFO is very much in line with the probabilistic programming spirit.

In the case where one pursues COFO via dynamic programming, it becomes {\it stochastic} dynamic programming because of the probabilistic sampling in the action.   The sampling step in the above can be specified in various ways, and incorporates the familiar (and familiarly tricky) exploration/exploitation tradeoff.   If probabilistic inference is used along with sampling, then one may have a peculiar sort  of approximate stochastic dynamic programming in which the step of choosing an action involves making an estimation that itself may be usefully carried out by stochastic dynamic programming (but with a different objective function than the objective function for whose optimization the action is being chosen).

Basically, in the COFO framework one looks at the process of optimizing $F$ as an explicit dynamical decision process conducted via sequential application of an operation in which: Operations $C_i$ that combine inputs chosen from a distribution induced by prior objective function evaluations, are used to get new candidate arguments to feed to $F$ for evaluation.   The reward function guiding this exploration is the quest for reduction of the entropy of the set of guesses at arguments that look promising to make $F$ near-optimal based on the evaluations made so far.

The same COFO process can be applied equally well the case of Pareto-optimizing a set of objective functions.   The definition of $M_\rho^D$ must be modified accordingly and then the rest follows.

Actually carrying out an explicit stochastic dynamic programming algorithm according to the lines described above, will prove computationally intractable in most realistic cases.   However, we shall see below that the formulation of the COFO process as dynamic programming (or simpler greedy sequential choice based optimization) provides a valuable foundation for theoretical analysis. 

\subsection{Cognitive Processes as COFO-Guided Metagraph Transformations} 

COFO is a highly general framework, and to use it to structure specific AI systems one has to take the next step and introduce specific sets of combinatory operations, often associated with specific incremental reward functions in the spirit of (but often not identical) the information-theoretic reward approach hinted above.   In \cite{goertzel2021patterns} explicit discussion is given to the COFO expression of a variety of cognitive algorithms used in the OpenCog AGI approach: Logical reasoning, evolutionary program learning, metagraph pattern mining, agglomerative clustering and activation-spreading-based attention allocation.

The use of distinction metagraphs as a general model of the processes underlying intelligence does not necessarily imply the use of metagraphs as a core concrete data structure in AGI system implementation.   Metagraphs could be used as a conceptual model of AI systems built using conventional neural net architectures, biologically realistic neural net simulations, topological quantum computing based AI fabrics (though here one might want quantum distinction graphs as discussed in \cite{Graphtropy}), or whatever.   However, it is a natural and in many ways compelling design choice to create an AGI architecture with metagraphs at the center -- thus placing metagraphs in a dual role: 

\begin{itemize}
\item As a fundamental means of analyzing what the AGI system is doing from a conceptual and phenomenological perspective
\item As the core data structure the AGI system uses to store various sorts of information as it goes about its business.
\end{itemize}

\noindent Obviously, the use of the same formal structure in these two roles makes it particularly straightforward for an AGI system to reflect on its own structure, to reason about its own operations and perceptions in a combined way and to intelligently guide self-modifications.    However, one could also certainly have advanced self-reflection and self-modification without the convenience of using the same mathematical structure on these two levels.

The series of proto-AGI architectures I have been involved in over the last decades -- Webmind, Novamente, OpenCog, Hyperon -- has been centered on the use of metagraphs in this dual role \footnote{In many of my prior writings I have referred to knowledge "hypergraphs" or "generalized hypergraphs" rather than "metagraphs"; but recently I decided that "metagraph" is less awkward terminologically than "generalized hypergraph" and made the switch.}.   In this sort of AGI architecture, the expression of logical inference, program learning and pattern mining in combinatory-system terms ties directly back to the discussion of distinction metagraphs and associated patterns in Section \ref{sec:ontology}, e.g.:

\begin{itemize}
\item   Logical inference rules can be considered as transformations on distinction metagraphs.   Bidirectional inference rules (expressed using coimplication) are rules mapping between two distinction metagraphs that have different surface form but ultimately express the same distinctions between the same observations
\item Programs can be viewed, using Curry-Howard type mappings, as series of steps for enacting these logical-inference-rule transformation on metagraphs, where the steps are to be carried out on an assumed reference machine.   The reference machine itself may also be represented as a distinction metagraph with temporal links used to express the transitions involved in computations.
\item Pattern mining (whether greedy or done via inference or say by evolutionary and/or probabilistic program learning methods) can be expressed in terms of patterns $y*z$ in metagraphs, where e.g. $z$ may be a distinction metagraph (perhaps a subset of a larger one) and $y$ may be a program enacting a transformation on distinction metagraphs (noting that this program may itself be expressed as a metagraph with causal links).   
\item Clustering can be viewed as a sort of metagraph transformation that creates new ConceptNodes grouping nodes into categories
\end{itemize}

In this context, COFO presents itself as a way of structuring processes via which sub-metagraphs transform other sub-metagraphs into yet other sub-metagraphs, where the submetagraphs are interpreted as combinators and are combined via a systematic recursive process toward the incremental increase of a particular reward function.

The common representation of multiple COFO processes involved in achieving the overall multiple-goal-achieving activities of a top-level DDS in terms of a shared typed metagraph is one way to facilitate the cognitive synergy needed to achieve high levels of general intelligence under practical resource constraints.   The reliance on a common metagraph representation makes it tractable for the multiple cognitive algorithms to share intermediate state as they pursue their optimization goals, which enables the cognitive-synergy dynamic in which each process is able to call on other processes in the system for assistance when it runs into trouble.

\subsection{COFO Processes as Galois Connections}  \label{sec:galois}

For some of the cognitive algorithms treated in COFO terms in  \cite{goertzel2021patterns} one requires a variety of COFO that uses greedy optimization to explore the dag of possibilities, for others one requires a variety of COFO that uses some variation on approximation stochastic dynamic programming.   In either case, one can use the "programming with Galois connections" approach from \cite{mu2012programming} to formalize the derivation of practical algorithmic approaches.   Roughly,  in all these cases, Galois connections are used to link search and optimization processes on directed metagraphs whose edge targets are labeled with probabilistic dependent types, and one can then show that -- under certain assumptions -- these connections are fulfilled by processes involving metagraph chronomorphisms (where a chronomorphism is a fold followed by an unfold, where both the fold and unfold are allowed to accumulate and propagate long-term memory as they proceed).

\subsubsection{Greedy Optimization as Folding}

Suppose we are concerned with maximizing a function $f:X \rightarrow R$  via a ``pattern search" approach.  That is, we assume an algorithm that repeatedly iterates a pattern search operation such as: Generates a set of candidate next-steps from its focus point $a$, evaluates the candidates, and then using the results of this evaluation, chooses a new focus point $a^*$.   Steepest ascent obviously has this format, but so do a variety of derivative-free optimization methods as reviewed e.g. in \cite{torczon1995pattern}.

Evolutionary optimization may be put in this framework if one shifts attention to a population-level function $f_P: X^N \rightarrow R$ where $X^N$ is a population of $N$ elements of $X$, and defines $f_P(x)$ for $x \in X^N$ as e.g. the average of $f(x)$ across $x \in X^N$ (so the average population fitness, in genetic algorithm terms).   The focus point $a$ is a population, which evolves into a new population $a^*$ via crossover or mutation -- a process that is then ongoingly iterated as outlined above.

The basic ideas to be presented here work for most any topological space $X$ but we are most interested in the case where $X$ is a metagraph.  In this case the pattern search iteration can be understood as a walk across the metagraph, moving from some initial position in the graph to another position, then another one, etc.

We can analyze this sort of optimization algorithm via the Greedy Theorem from \cite{mu2012programming}, 

\begin{thm}
(Theorem 1 from \cite{mu2012programming}). $\llparenthesis S \upharpoonright R \rrparenthesis \subseteq \llparenthesis S \rrparenthesis \upharpoonright R$ if R is transitive and S satisfies the "monotonicity condition" $R^\circ \leftarrow{S} FR^\circ$
\end{thm} \label{thm:Greedy}

\noindent which leverages a variety of idiosyncratic notation:

\begin{itemize}
\item $R \xleftarrow{S} FR$ \textrm{ indicates } $S \cdot FR \subseteq R \cdot S$
\item $\llparenthesis S \rrparenthesis$ \textrm{ means } the operation of folding $S$
\item $\langle \mu X :: f X \rangle$ denotes the least fixed point of $f$
\item $T^\circ$  \textrm{ means } the converse of $T$, i.e. $(b,a) \in R^\circ \equiv (a,c) \in R$
\item $S \upharpoonright R$  \textrm{ means } "$S$ shrunk by $R$", i.e. $S \cap R / S^\circ$
\end{itemize}

\noindent Here $S$ represents the local candidate-generation operation used in the pattern-search optimization algorithm, and $R$ represents the operation of evaluating a candidate point in $X$ according to the objective function being optimized.   

If the objective function is not convex, then the theorem does not hold, but the greedy pattern-search optimization may still be valuable in a heuristic sense.   This is the case, for instance, in nearly all real-world applications of evolutionary programming, steepest ascent or classical derivative-free optimization methods.

\subsubsection{Galois Connection Representations of Dynamic Programming Decision Systems Involving Mutually Associative Combinatory Operations} \label{sec:Galois}

Next we consider how to represent dynamic programming based execution of DDSs using folds and unfolds.  Here our approach is to leverage Theorem 2 in \cite{mu2012programming} which is stated as

\begin{thm}
(Theorem 2 from \cite{mu2012programming}). Assume $S$ is monotonic with respect to $R$, that is, $R \xleftarrow{S} F_R$ holds, and $dom(T) \subseteq dom(S \cdotp FM)$.   Then 

$$
M=(  \llparenthesis S \rrparenthesis \cdotp  \llparenthesis T \rrparenthesis^\circ ) \upharpoonright R \Rightarrow \langle \mu X :: (S \cdotp FX \cdotp T^\circ) \upharpoonright  R \rangle \subseteq M
$$

\end{thm} \label{thm:DP}

Conceptually, $T^\circ$ transforms input into subproblems, e.g.

\begin{itemize}
\item for backward chaining inference, chooses $(x,y,C)$ so that $z = C(x,y)$ has high quality (e.g. CWIG)
\item for forward chaining, chooses x, y, C so that z = C(x,y) has high interestingness (e.g. CWIG)
\end{itemize}

\noindent $FX$ figures out recursively which combinations give maximum immediate reward according to the relevant measure.   These optimal solutions are combined and then the best one is picked by $\upharpoonright R$, which is the evaluation on the objective function.   Caching results to avoid overlap may be important here in practice (and is what will give us histomorphisms and futumorphisms instead of simple folds and unfolds).

The fix-point based recursion/iteration specified by the theorem can of course be approximatively rather than precisely solved -- and doing this approximation via statistical sampling yields stochastic dynamic programming.  Roughly speaking the approach symbolized by $M=(  \llparenthesis S \rrparenthesis \cdotp  \llparenthesis T \rrparenthesis^\circ ) \upharpoonright R $  begins by applying all the combinatory operations to achieve a large body of combinations-of-combinations-of-combinations-$\ldots$, and then shrinks this via the process of optimality evaluation.  On the other hand, the least-fixed-point version on the rhs of the Theorem iterates through the combination process step by step (executing the fold).

\subsection{Associativity of Combinatory Operations Enables Representing Cognitive Operations as Folding and Unfolding}

A key insight reported in {\it Patterns of Cognition} is that the mutual associativity of the combinatory operations involved in a cognitive process often plays a key role in enabling the decomposition of the process into folding and unfolding operations.   This manifests itself for example in the result that

\begin{thm}
A COFO decision process whose combinatory operations $C_i$ are mutually associative can be implemented as a chronomorphism.
\end{thm}

This general conclusion regarding mutual associativity resonates fascinatingly with the result from \cite{goertzel2020grounding} mentioned above, that mutually associative combinatory operations lead straightforwardly to subpattern hierarchies.   We thus see a common mathematical property leading to elegant and practically valuable symmetries in both algorithmic dynamics and in knowledge-representation structure.     This bolsters confidence that the combinatory computational model is a good approach for exploring the scaling-down of generic but infeasible AGI models toward the realm of practically usable algorithms.

This conclusion regarding mutual associativity also has some practical implications for the particulars of cognitive processes such as logical reasoning and evolutionary learning.   For instance, one can see that mutually associativity holds among logical inference rules if one makes use of reversible logic rules (co-implications rather than implications), and for program execution processes if one makes use of reversible computing.    It is also observed that where this mutual associativity holds, there is an alignment between the hierarchy of subgoals used in recursive decision process execution and subpattern hierarchies among patterns represented in the associated knowledge metagraph.

In the PLN inference context, for example, the approach to PLN inference using relaxation rather than chaining outlined in \cite{Goertzel2008prob} is one way of finding the fixed point of the recursion associated with the COFO process.    What the theorem suggests is that folding PLN  inferences across the knowledge metagraph is another way, basically boiling down to forward and backward chaining as outlined above.  However, it seems this can only work reasonably cleanly for crisp inference if mutual associativity among inference rules holds, which appears  to be the case only if one uses PLN rules formulated as co-implications rather than one-way implications.

Further, when dealing with the uncertainty-management aspects of PLN rules, one is no longer guaranteed associativity merely by adopting reversibility of individual inference steps, and one is left with a familiar sort of heuristic reasoning: One tries to arrange one's inferences as series of co-implications whose associated distributions have favorable independence relationships.   For instance if one is trying to fold forward through a series of probabilistically labeled co-implications, one will do well if each co-implication is independent of its ancestors conditional on its parents (as in a Bayes net); this allows one to place the parentheses in the same place the fold naturally does.   The ability of chronomorphisms to fulfill the specifications implicit in the relevant Galois connectiosn becomes merely an approximate heuristic guide, though we suspect still a very valuable one.

\subsection{The Challenge of Handling Dynamic Knowledge Base Revisions }

The assumptions needed to get from the symmetry properties of discrete decision processes to fold and unfold operations are not entirely realistic -- for instance, to get the derivations to work in their most straightforward form, one needs to assume the underlying metagraph remains unchanged as the folding and unfolding processes proceed.   If the metagraph changes dynamically along with the folding and unfolding -- e.g. because inference processes are drawing conclusions from the nodes and links created during the folding process, and these conclusions are being placed into the metagraph concurrently with the folding process proceeding -- then one loses the straightforward result that simple approximate stochastic dynamic programming algorithms will approximate the optimal result of the decision process.   

There may be more sophisticated, conceptually similar results that do apply in this case, but these have yet to be worked out.   This is a serious limitation, but it must also be understood that in many cases the real-time changes to the metagraph incurred by the folding and unfolding process are not a significant factor.   Creating rigorous theory connecting abstract AGI theory to pragmatically relevant cognitive algorithms and their implementations is a complex matter that will require elaboration of a variety of interrelated subtheories; the big question is when this process of derivation will be taken over by the AI itself.   The process of derivation of algorithms from operator symmetries as represented by Galois connections is intentionally ideally designed for ultimate execution by automated theorem-provers rather than people.

\subsection{The Relation Between Maximally-General AGI and Specific Useful Algorithms}

Via exploiting particular algebraic properties of combinatory systems such as mutual associativity, and using these to drive derivation of particular algorithmic approaches like metagraph chronomorphism, we are moving in this {\it Patterns of Cognition}  work from high level generalities about maximally-general-scope general intelligence to reasonably more specific conclusions about specific algorithms.   From a maximally-general view one is thus shifting focus to AGI systems that are biased toward specific sorts of goals and specific sorts of environments -- e.g. those in which evolutionary operators for program evolution will tend to work better than brute force search of program space, those in which agglomerative clustering will tend to work better than brute force search over all data partitions, those in which intensional and extensional reasoning are often well correlated and suitable to guide each other, etc.   

If one is interested in algorithmic performance as averaged over all computable goals and all computable environments, then specific algorithms like evolutionary program learning and agglomerative clustering, or specific inference approaches like intensional reasoning, are likely of very limited interest -- because while there are some goals and environments for which they work nicely, there are also (theoretically and mathematically speaking) a lot of goals and environments for which they are counterproductive and just waste extra time relative to brute force methods.   However, if one is interested a bit more narrowly in performance over goals and environments that are relevant specifically to humans or humans' near term reactions -- or that are relevant in our physical universe rather than on average across all computationally specifiable physical universes -- then these specific algorithms may indeed be highly relevant.  And it is then interesting to see what one can figure out about effective implementation of these algorithms via looking at how their underlying formal symmetry properties enable specialization of maximally-general-purpose AGI approaches.

\section{Critical Priors for Human-Like or Human-Friendly General Intelligence} \label{sec:human-priors}

We have suggested that concision in a combinatory computational model, perhaps one represented as a CoDD, and one involving sets of mutually associative and cost-associative combinatory operations, gives a prior distribution of significant relevance to general intelligence.   We have argued that the cognitive algorithms utilized in OpenCog have relatively simple expression in such a computational model.   It is also interesting, however, to look at prior distributions valuable for guiding human-like AGI in more everyday-life-oriented way.

Yoshua Bengio and colleagues attracted attention in 2017 for a paper on what they called the "Consciousness Prior" \cite{bengio2017consciousness} -- a prior distribution over environments and goals matching the particularities of human intelligence.   I think consciousness is a fascinating issue (and will discuss it in Section \ref{sec:cons} below) but that giving  consciousness a prime focus when talking about prior distributions is unnecessary and likely to cause confusion.   My own take in this direction, published in 2009, was called the "Embodied Communication Prior" \cite{goertzel2009embodied}.   Of course consciousness is richly tied up with embodied communication, but it seems more concrete progress can be made in this context by focusing on the embodied communication process rather than on consciousness per se.

The core idea Bengio presents that AGI needs a prior distribution that favors joint distributions that factor into forms where most weight goes to a small number of factors.   This is a very sensible idea and does indeed tie in with the way working memory works in current human and AI minds.    Put together with the assumption of a combinatory computational model and mutual associativity among many combinations, this certainly considerably narrows down the class of prior distributions.   However, if closely human-like general intelligence is what one's after, I don't think this sort of general mathematical analysis goes far enough.

It would appear that the structure and dynamics of human-like minds have been adapted heavily to considerably more specialized assumptions to do with modeling events in 4D spacetime, and specifically to handling communication among spatiotemporally embodied agents who share the same sensation and action space.   In the paper {\it A Mind-World Correspondence Principle} \cite{goertzel2013mind} I sought to couch this sort of adaptation in a very general way.   I suggest that symmetries and other regularities in the environments and goals that an intelligence needs to deal with, should be mappable via (uncertain) morphisms into corresponding symmetries/regularities in the structure and dynamics of the intelligent system itself; in the paper I roughly formalize this correspondence in terms of category theory.

E.g. one feature of the environments and goals human-like minds are faced with, is that they tend to factorize into qualitatively different types of knowledge / perception / action -- e.g. procedural vs. declarative/semantic vs. attentional vs. sensory, etc.   The "mind-world correspondence here" lies in the waysdifferent types of human memory as recognized by cognitive psychology correspond closely to different aspects of the everyday human physical and social environment. 

Different types of knowledge tend to require different sorts of learning algorithm, which leads one beyond the complex issues of translation between knowledge types, and into the yet more complex issues of translation between the different learning algorithms corresponding to different types of knowledge, including their intermediate states as well as their final conclusions.   This line of thinking leads to the hypothesis of minds that have distinct yet closely coupled subcomponents that need to have robust capability to help each other out of difficult cognitive spots -- i.e. "Cognitive Synergy".
\subsection{Formalizing Cognitive Synergy}

The paper {\it Toward a Formal Model of Cognitive Synergy} \cite{DBLP:journals/corr/Goertzel17} develops a formal characterization of "cognitive synergy" between cognitive processes making decisions based on iterative metagraph sampling (what is called there PGMC, Probabilistic Mining and Growth of Combinations, a conceptually similar predecessor to the COFO formulation given above).   The characterization is based on developing a formal notion of "stuckness," and defining synergy as a relationship between cognitive processes in which they can help each other out when they get stuck. It is proposed that cognitive processes relating to each other synergetically, associate in a certain way with functors that map into each other via natural transformations. Cognitive synergy is proposed to correspond to a certain inequality regarding the relative costs of different paths through certain commutation diagrams.

The related paper {\it Symbol Grounding via Chaining of Morphisms} \cite{lian2017symbol} explains how the connection between language, action, perception and memory works in terms of this category-theoretic model of cognitive synergy.   Explicit examples are given showing elegant morphisms between natural language syntax, the compositional grammar of actions and perceptions, and the logical structure of inferences.   Taking these morphisms that exist on the knowledge representation level, and mapping them into the learning algorithms corresponding to the different sorts of knowledge representation, one then obtains the cross-cognitive-process morphisms that constitute cognitive synergy.

Part of the conceptual upshot of this sort of model is that, ultimately, the cognitive processing of real-world AGI systems can be viewed as: a set of interacting cognitive algorithms, each of which in a sense results from doing program specialization on the universal algorithm "form an algorithmic Markov model consistent with one's observations, and use it to drive inference about what procedures will achieve one's goals given the observed context", relative to focus on a specific sort of knowledge, memory or situation (e.g. procedural, sensory, declarative...).   These specialized cognitive algorithms must be learned/evolved based on multiple constraints including energetic usage, minimizing spatial extent and maximizing processing speed, and interoperability among the different cognitive algorithms (so that they can see each others' internal states so as to help each other out when they get stuck).

Design of an AGI framework like OpenCog  may be viewed as performing this sort of program specialization "by hand", as we don't have automated program specializers capable of this degree of complexity.  An AGI program specializer will be able to do it, but then we have a chicken-egg problem -- which is solved by human AGI system designers performing the first round of the iteration.   Techniques like automatic derivation of algorithms from specifications using Galois connections represent initial steps toward making this sort of AGI program specialization feasible, by using pragmatically motivated prior assumptions about the relevant program space to constrain  and guide the search.

\subsection{Cognitive Architecture of Human-Like Minds} \label{sec:cogarch}

\begin{figure}[htb]
\centering
\includegraphics[width=12cm]{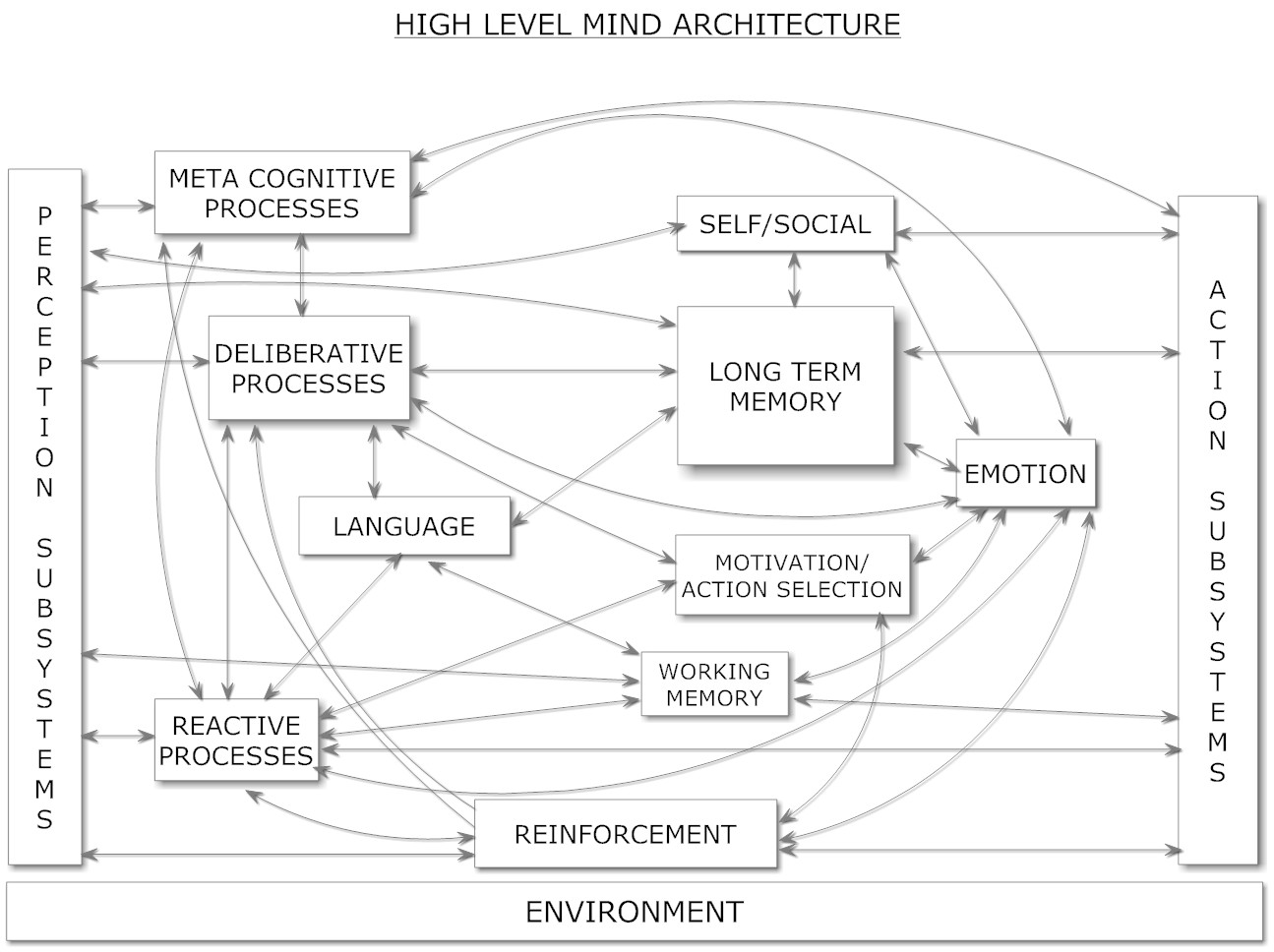}
\caption{High level architecture diagram for a human-like general intelligence, inspired by the work of Aaron Sloman among other sources.  This can be viewed in an obvious way as a particular way of refining Figure \cite{ref:dds}.  From \cite{Goertzel2012aa}}
\label{fig:mind_arch_sloman}
\end{figure}

\begin{figure}[htb]
\centering
\includegraphics[width=12cm]{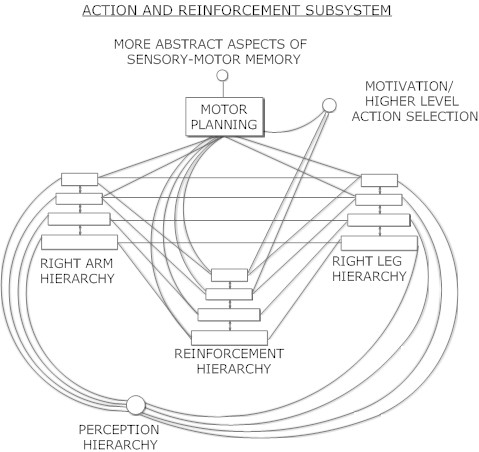}
\caption{High level architecture diagram for the subnetwork of a human-like general intelligence focused on action.  From \cite{Goertzel2012aa}}
\label{fig:mind_arch_action}
\end{figure}

\begin{figure}[htb]
\centering
\includegraphics[width=12cm]{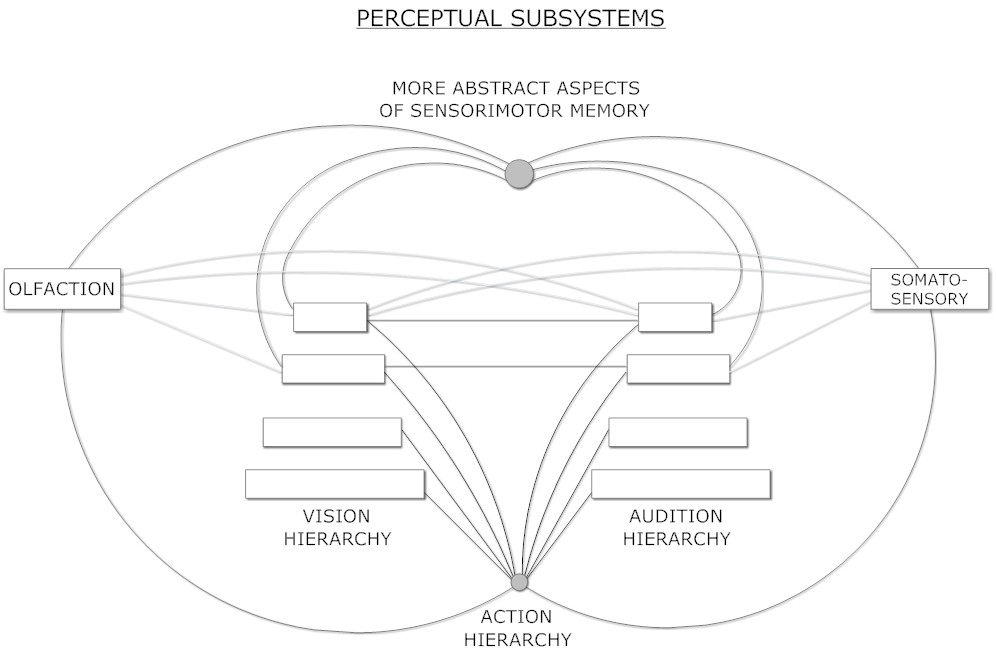}
\caption{High level architecture diagram for the subnetwork of a human-like general intelligence focused on perception  From \cite{Goertzel2012aa}}
\label{fig:mind_arch_perception}
\end{figure}

\begin{figure}[htb]
\centering
\includegraphics[width=12cm]{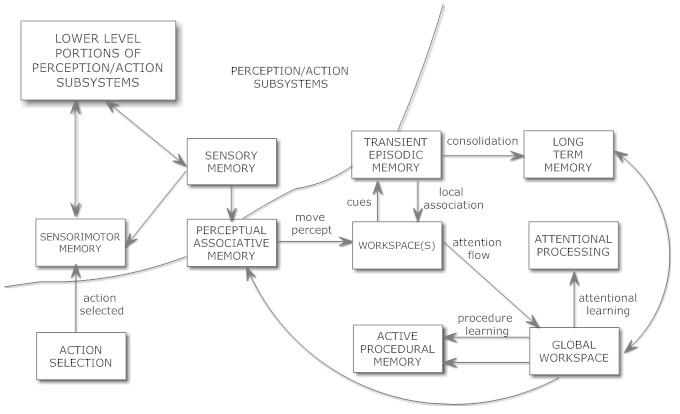}
\caption{High level architecture diagram for the subnetwork of a human-like general intelligence concerned centrally with working memory.   Inspired by the work of Stan Franklin on the LIDA cognitive architecture, among other sources.  From \cite{Goertzel2012aa}}
\label{fig:mind_arch_LIDA}
\end{figure}

\begin{figure}[htb]
\centering
\includegraphics[width=12cm]{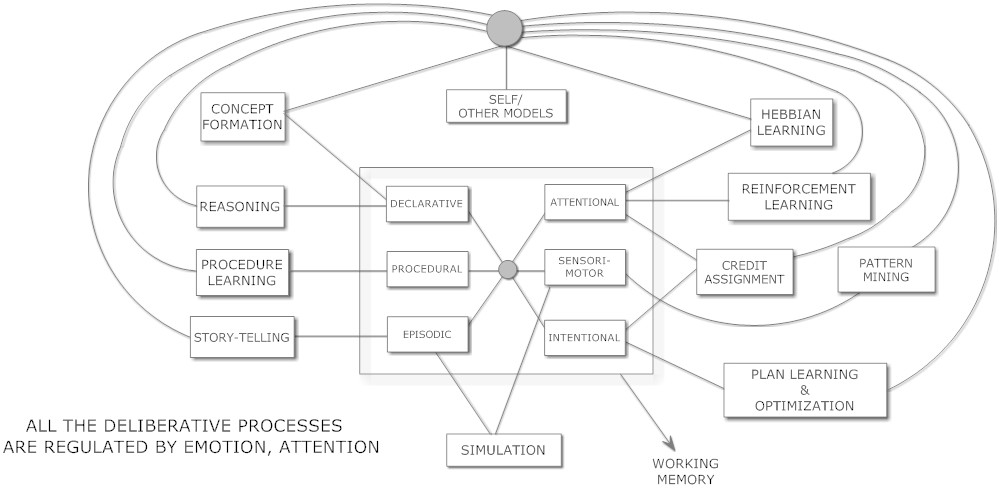}
\caption{High level architecture diagram for the subnetwork of a human-like general intelligence focused on long-term memory and closely associated reasoning and learning processes.  From \cite{Goertzel2012aa}}
\label{fig:mind_arch_LTM}
\end{figure}

\begin{figure}[htb]
\centering
\includegraphics[width=12cm]{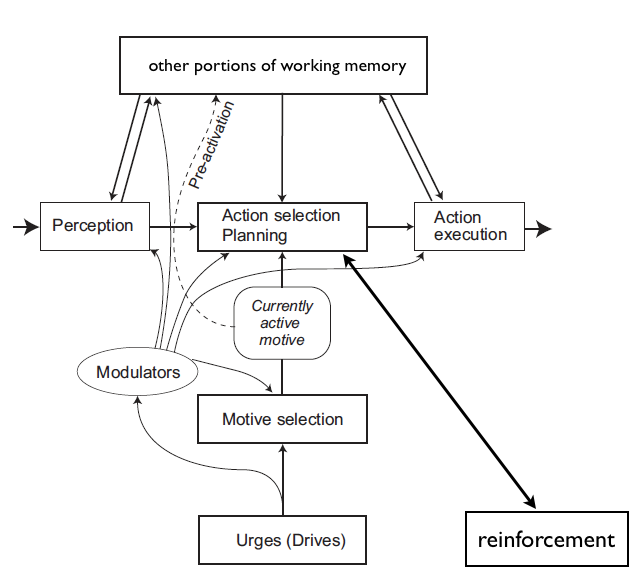}
\caption{High level architecture diagram for the subnetwork of a human-like general intelligence focused on motivation.  Inspired by the work of Joscha Bach on the Psi cognitive model along with other sources.   From \cite{Goertzel2012aa}}
\label{fig:mind_arch_psi}
\end{figure}

\begin{figure}[htb]
\centering
\includegraphics[width=12cm]{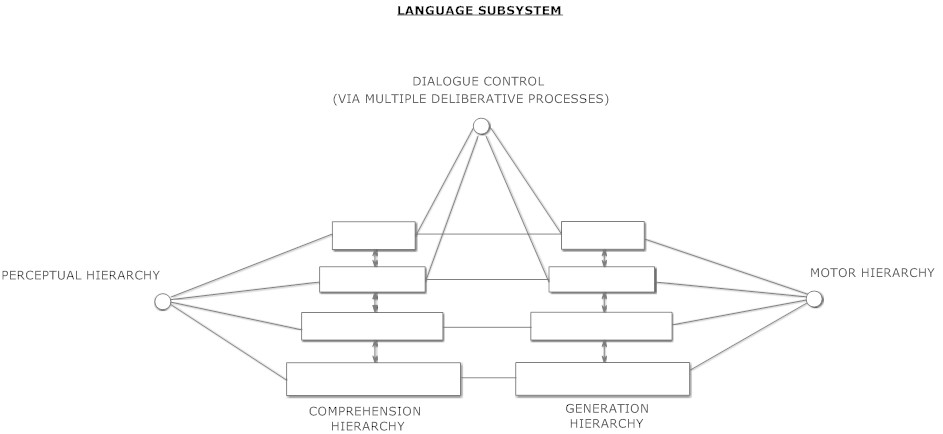}
\caption{High level architecture diagram for the subnetwork of a human-like general intelligence focused on natural language processing.  From \cite{Goertzel2012aa}}
\label{fig:mind_arch_language}
\end{figure}

In \cite{Goertzel2012aa} Matt Ikle' and I synthesized theories of the cognitive architecture of human and human-like minds from a number of different AI theorists and cognitive scientists, arriving at a set of diagrams shown in Figures \ref{fig:mind_arch_sloman} - \ref{fig:mind_arch_language}, encapsulating the key components and interactions needed to realize human-like intelligence according to the best current knowledge of cognitive science as integrated by a number of deep-thinking AGI theorists.   Conceptually, one can see Figure \ref{fig:mind_arch_sloman} as a refinement of the DDS model as shown in Figure \ref{fig:dds}, dividing the perception / cognition / memory box from the generic DDS model into a number of sub-units reflecting the modularization enforced by human-like embodiment.   The distinction between working memory and long term memory, or the distinction between perception and deliberation, for instance, are not necessary in the general DDS model but make obvious sense in the context of intelligent systems dealing with human-relevant goals in everyday human environment under relatively strict processing constraints.   The other diagrams in the series expand the boxes in Figure  \ref{fig:mind_arch_sloman}  into subprocesses in a way also reflecting the intersection of human-relevant goals, environments and processing constraints.

One weakness of diagrams of this nature, of course, is that they do not explicate the nature of the interactions between the components, nor the representations nor learning algorithms used within the components -- which in the end are much deeper matters than drawing a bunch of boxes and lines.   The theory of cognitive synergy, and associated theoretical notions reviewed above, mostly deal with these deeper matters.  However, it's also important and interesting to see how these representations, dynamics and interactions manifest themselves in terms of high level components and information flows.

The cognitive processes associated with long-term memory, as summarized in Figure \ref{fig:mind_arch_LTM}, are where most of my effort as an AGI researcher has been spent, and what occupies most of the 900+ pages of my core AGI work {\it Engineering General Intelligence} ({\it EGI}).   {\it EGI} considers cognitive mechanisms such as declarative knowledge creation via probabilistic reasoning, procedure learning via probabilistic evolutionary program learning, pattern detection via greedy hypergraph pattern mining, and attention allocation via artificial economics.   Each of these is viewed in terms of combinatory systems involving atomic entities represented in a dynamic knowledge metagraph.   The treatment of these various cognitive processes in terms of COFO and Galois connections given in \cite{goertzel2021patterns} is more sophisticated than the presentation in  {\it EGI} but is very much in the same spirit.  The reasons for having these particular sorts of COFO processes within one's AGI DDS are complex and have to do partly with mathematical considerations, and partly with the specifics of achieving human-relevant goals in human-relevant environments using strictly limited resources, a matter discussed at length in {\it EGI}  from a variety of perspectives.

\section{Situating the OpenCog Hyperon Design in General AGI Theory} \label{sec:hyperon}

The theoretical ideas presented here could be manifested in practical AGI designs in multiple ways; the approach which I'm currently pursuing is OpenCog Hyperon, a significantly improved and upgraded new version of the OpenCog AGI architecture, with a high-level structure as shown in Figure \ref{fig:hyperon}.   Hyperon's design has the following key aspects:

\begin{figure}[htb]
\centering
\includegraphics[width=12cm]{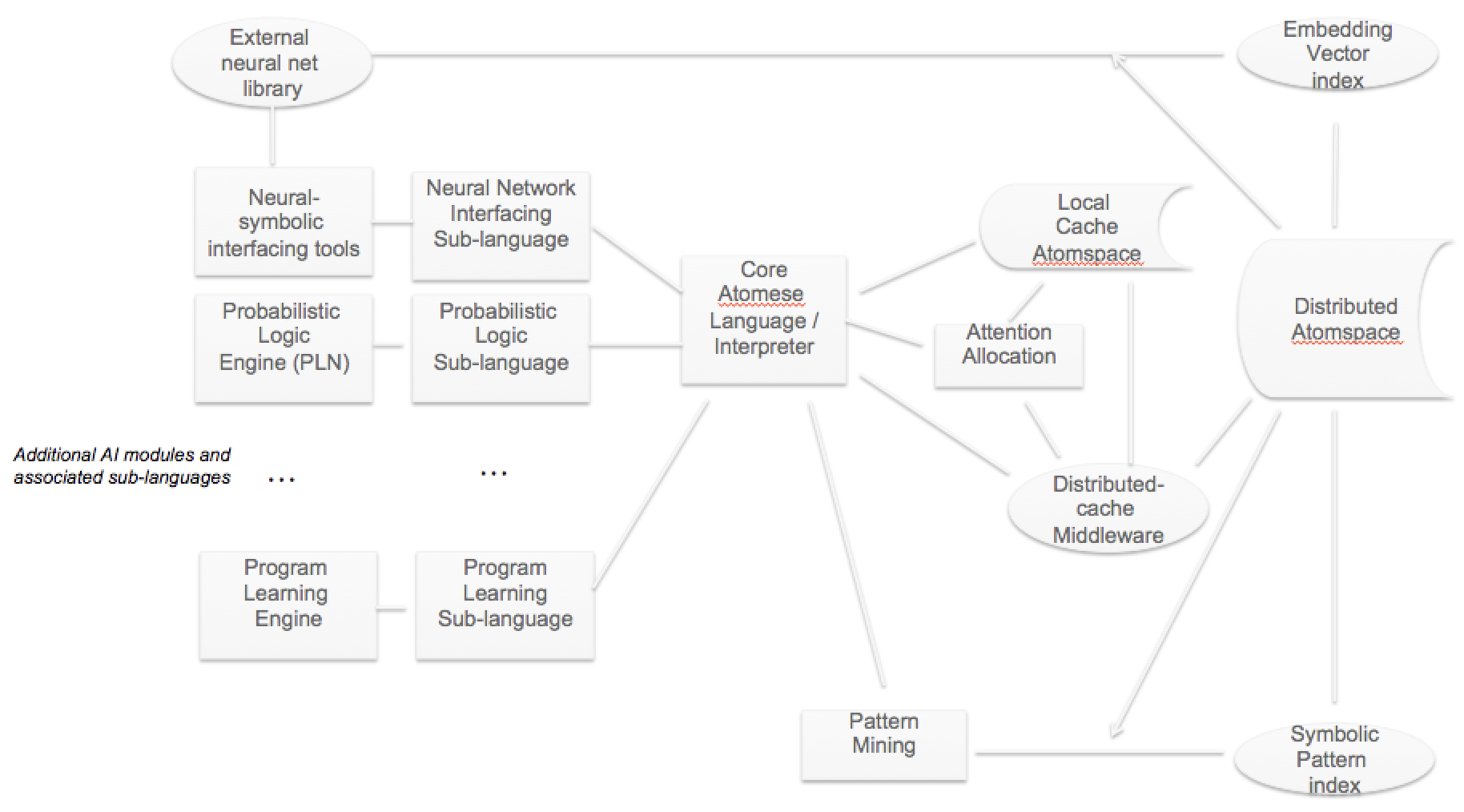}
\caption{Preliminary high level architecture of OpenCog Hyperon system.}
\label{fig:hyperon}
\end{figure}

\begin{itemize}
\item A scalable, distributed (and in some deployments decentralized) metagraph knowledge store called the Atomspace (as noted above, in Hyperon and its predecessor architectures one uses metagraphs as a core concrete meta-representational fabric as well as in the role of a higher-level descriptive language)
\item A control framework for enacting multiple AI algorithms that interact with the Atomspace, centrally including algorithms that operate according to the DDS and COFO methodology
\item A framework for controlling an intelligent agent (which could be a robot, an avatar, or a system with a software interface) using a DDS process driven by multiple goals organized according to the variant of the Psi model shown in Figure \ref{fig:mind_arch_psi} as discussed in \cite{goertzel2020grounding}
\item A system of multiple COFO processes that are organized to learn implications of the form $\textrm{Context } \& \textrm{ Procedure} \rightarrow \textrm{Goal}$, where 
\begin{itemize}
\item $\textrm{Context }$ is a pattern in the system's observations (potentially including observations of itself), represented in the Atomspace metagraph
\item $\textrm{ Procedure}$ is an executable procedure, implemented in the Atomspace metagraph
\item The $\textrm{Goal}$s are the goals of the agent's top-level DDS 
\end{itemize}
\end{itemize}

\noindent The COFO processes in the Hyperon design are based on a handful of core cognitive algorithms as described in \cite{EGI1} \cite{EGI2}: Probabilistic Logic Networks (PLN) reasoning, probabilistic evolutionary program learning, pattern mining, agglomerative clustering, association-spreading-based attention-allocation etc.  The general OpenCog conceptual AGI design -- which is common to the original OpenCog system and the new Hyperon version -- explains how the Atomspace metagraph and these cognitive algorithms fill all the roles specified in the human-like-cognitive-architecture diagrams given in Section \ref{sec:cogarch}.   A great deal of the discussion in  \cite{EGI1} \cite{EGI2} is about explaining exactly how these algorithms, implemented on a common knowledge metagraph, can be made to fulfill these cognitive roles.   

The "cognitive module network" design pattern articulated by Alexey Potapov and colleagues in \cite{potapov2019cognitive} in 2019 (a more recent development than the treatment of similar themes in \cite{EGI1} \cite{EGI2} which were published in 2014 ) is also used in Hyperon to integrate deep neural networks and other external scalable floating-point-vector-oriented AI learning algorithms tightly with metagraph-based COFO processes.  Some of the improvements made in Hyperon compared to the original OpenCog system are oriented toward increasing the simplicity and efficiency of this cognitive-module-network style integration.   From a high level theoretical perspective this is just about decreasing the frictional cost between different COFO processes integrated as subordinates to the high level DDS controlling overall system activity.

The explication of how the various aspects of human-like cognitive function can be carried out via application of a small number of DDS and COFO processes acting on a common metagraph, arranged and interoperating in various ways and manifesting cognitive synergy, becomes a lengthy and intricate story, which is very briefly summarized in Section \ref{sec:humancog} below.   From a high level AGI theory perspective, though, what this is about is manifesting the general DDS / COFO / distinction-metagraph approach outlined above in the context of the specific environmental biases and resource constraints characterizing human-like intelligence.

\subsection{Achieving Human-Like Cognitive Processes via DDS and COFO} \label{sec:humancog}

Without replicating the nearly-1000 page treatment given in \cite{EGI1} \cite{EGI2}, we will here give a brief run-through of how the key aspects of human-like cognition summarized above fit into the DDS/COFO-on-typed-metagraphs framework we have sketched above.   We will occasionally lapse here into OpenCog lingo, e.g. using the term Atom to encompass either nodes or links within metagraphs.

\begin{itemize}
\item {\bf Motivational Subsystems} (Fig. \ref{fig:mind_arch_psi})
\begin{itemize}
\item {\bf Action selection} is carried out via probabilistic selection of procedures based on $\textrm{Context } \& \textrm{ Procedure} \rightarrow \textrm{Goal}$ triads, which may be modeled as the action-choice portion of the top-level DDS in the overall cognitive architecture
\item {\bf Planning} is procedure learning resulting in creation of $\textrm{Context } \& \textrm{ Procedure} \rightarrow \textrm{Goal}$ triads where the procedures involve plans.   Plan learning may emerge in many ways, e.g. directly by automated program learning that produces plan-embodying procedures, or via automated reasoning that produces declaratively-specified plans that are then automatically transformed into executable procedures.
\item {\bf Action execution} involves executing the selected procedures, which may involve calls into other cognitive processes if the procedures involve component sub-actions that are abstractly defined, and may also involve invocation of neural networks or other Atomspace-external software processes (according to the cognitive-module-networks design pattern)
\item {\bf Motive selection}, the selection of which of the system's potentially multiple top-level goals to focus on at a given point in time, is reflected by the adjustment of {\it short term importance} (STI) values associated with the Atoms representing goals.   These STI values will be updated via importance-spreading dynamics (according to the equations of ECAN, Economic Attention Networks that spread importance among all Atoms in Atomspace), or in an advanced system may also be adjusted based on inference or other cognitive processes.
\item {\bf Basic drives and urges}, on which the system's top-level goals are based, are initially supplied by the human creators of the system, though ultimately may (and arguably should) be susceptible to self-modification.   For instance if there is a top-level goal to keep the amount of novelty experienced during each time-interval within a certain range, there will be a basic "drive" toward novelty underlying this, represented by a procedure that evaluates the degree of novelty experienced during a time-interval.
\item {\bf Reinforcement} involves updating of the truth values of the implication links $\rightarrow$ in implications of the form  $\textrm{Context } \& \textrm{ Procedure} \rightarrow \textrm{Goal}$.   These truth values may be updated by a variety of mechanisms, including directly based on experience, indirectly based on probabilistic inference or pattern mining, or even e.g. by backpropagation through the Atomspace.
\end{itemize}
\item {\bf Action and Reinforcement Subsystems} (Fig. \ref{fig:mind_arch_action})
\begin{itemize}
\item {\bf Motor movement planning} is a mix of program learning, probabilistic reasoning, and neural learning accessed from Atomspace via cognitive modules
\item {\bf Motor movement hierarchies} are represented in Atomspace via subpattern hierarchies representing hierarchical patterns of movement, morphically mapped into procedural hierarchies comprising routines decomposed into subroutines
\item The role of {\bf reinforcement hierarchies} in motor learning is played by multiple processes cooperating to update the implication links in the subpattern hierarchies mapping into executable procedure hierarchies embodying motor movements
\end{itemize}
\item {\bf Perceptual Subsystems} (Fig. \ref{fig:mind_arch_perception})
\begin{itemize}
\item {\bf Perceptual hierarchies } (e.g. vision and audition)  are represented as subpattern hierarchies whose upper levels comprise procedures explicitly represented in the Atomspace, and whose lower levels comprise neural modules symbolized in and accessed from the Atomspace
\item {\bf Attractor-based perception} (e.g. human olfaction) is manifested by richly interconnected networks of Atoms representing low-level perceptual patterns, with nonlinear importance (STI) spreading dynamics creating attractors.   The Atoms in these networks will sometimes themselves refer to attractor or transient patterns in attractor neural nets external to Atomspace, yielding a multilevel neural-symbolic attractor network.
\item {\bf Perception-action feedback} is achieved via spread of importance between perception and action oriented Atoms, and also by Atoms explicitly denoting formal relationships enacting e.g. program transformations relating perceptual procedures and movement procedures, and Curry-Howard type transformations between e.g. executable procedures connoting physical actions and declarative patterns relating perceptions or between perceptual procedures and declarative patterns describing movements.
\end{itemize}
\item {\bf Working Memory Centric Subsystems} (Fig. \ref{fig:mind_arch_LIDA})
\begin{itemize}
\item {\bf Sensorimotor, sensory and motor memory} is a mix of patterns in neural nets referenced by the Atomspace, and more abstract symbolic patterns with probabilistic semantics combining these lower-level neural patterns
\item {\bf Perceptual associative memory} comprises links connecting perceptual patterns with other Atoms, with basic associative semantics according to importance spreading and more abstract semantics in accordance with the types of the links
\item {\bf Transient episodic memory} is comprised by links representing recent experience perceptually, conceptually and action-wise -- with high Short-Term Importance and low Long-Term Importance, by default
\item {\bf Attentional processing} is the activity of the ECAN dynamics regulating importance values, which among other aspects maintains the "moving bubble of attention"
\item {\bf Global workspace} is an aspect of attentional dynamics wherein a variety of Atoms in the Atomspace have their importance stimulated due to their direct or indirect linkage to the Atoms with currently high STI value
\item {\bf Local workspaces} are collections of tightly interlinked Atoms that achieve high STI value for a period of time due to their common engagement with a certain cognitive process.    Their interlinkage may rapidly increase due to this common engagement.  In a distributed implementation a local workspace concerned with a certain sort of processing (e.g. language processing or robot movement) may also be physically localized for efficiency's sake.
\end{itemize}
\item {\bf Long Term Memory Centric Subsystems} (Fig. \ref{fig:mind_arch_LTM})
\begin{itemize}
\item {\bf Deliberative reasoning} is centrally carried out via Probabilistic Logic Networks based reasoning with experience-guided adaptive inference control
\item {\bf Procedure learning} is centrally carried out via probabilistic evolutionary program learning (i.e. Atomspace-MOSES, in OpenCog lingo), augmented by neural net learning methods for lower-level procedures invoked as subroutines via Atomspace-embedded programs
\item {\bf World Simulation} is embodied in declarative probabilistic implications expressing constraints used to guide the operation of external simulation engines that are referenced in Atomspace as cognitive modules
\item {\bf Self-modeling}, {\bf Episodic memory}  and {\bf Story-telling} are high level emergent functions involving probabilistic inference, procedure learning, attention allocation and concept formation coordinated together, invoking lower-level neural net functionality as needed
\item {\bf Concept Formation} is executed by multiple heuristics creating new Atoms from old, e.g. concept blending as explored in \cite{EGI2} and paraconsistent formal concept analysis as suggested in \cite{goertzel2020paraconsistent}
\item {\bf Pattern Recognition} and {\bf Pattern Mining} are embodied in search processes that scan Atomspace for significant patterns according to variants of the formal definition of pattern reviewed above, and create new Atoms embodying these patterns
\item {\bf Credit Assignment} is embodied in the creation of causal links joining Atoms representing internal actions and Atoms representing results -- which may be formed via explicit probabilistic causal reasoning, or in appropriate cases via subsymbolic reinforcement learning methods configured to utilize appropriate probabilistic semantics.  
\item {\bf Plan learning} is carried out in cognitive contexts similarly to in motor movement contexts, via a combination of declarative PLN inference and probabilistic evolutionary program learning, with the option to follow up abstract plan learning with low-level plan optimization carried out via external heuristics
\end{itemize}
\item {\bf Language} (Fig. \ref{fig:mind_arch_language})
\begin{itemize} 
\item Language {\bf comprehension} and  {\bf generation} are viewed in Hyperon as particular cases of perception and action processing and are handled at the high level via the same combination of cognitive processes.   However, the strategy prototyped in \cite{goertzel2020guiding} may be used to allow large-scale neural language models to serve as oracles guiding more abstract conceptual language learning processes.
\item {\bf Dialogue control} is viewed in Hyperon as a particular case of motivated action, with "speech acts" best considered as interleaved with other sorts of acts chosen by the top-level DDS when appropriate for achieving top-level goals \cite{goertzel2010general}.
\end{itemize}
\end{itemize}

Clearly, architecting an AGI system containing all these components is a substantial undertaking and the particulars can be worked out in various ways, depending among other factors on the practical interfaces and applications one has in mind.   However, the key point from the present perspective is: A practical AGI design corresponding to human-like cognitive architecture can be created via taking the key processes identified by cognitive science, realizing each one via a DDS (in many cases a COFO DDS), and then connecting these DDSs appropriate via a shared knowledge metagraph -- where the metagraph is used both to implement the procedures and rewards underlying the DDS, and as a shared repository for background knowledge and intermediate state, on which cross-DDS attention-allocation DDSs operate and modulate the presentation of background knowledge to the DDSs in the overall system.   This is the crux of the OpenCog approach historically, currently under refinement into the OpenCog Hyperon design.

\subsection{Theoretical Guidance for AGI Programming Language Design}

Implementing the above in a practical metagraph/DDS/COFO based software system obviously involves a great number of fine-grained design choices, some of which have been made differently throughout the series of proto-AGI architecture serving as predecessors to our current work on OpenCog Hyperon.   Many of the key choices we currently face in Hyperon design have been encapsulated in the task of designing the Atomese 2 programming language, which on the back end takes the form of a particular system of dependent types used to label the nodes and edges in Atomspace metagraphs (and on the front end will be a syntactically-sugared way allowing humans to specify and read Atomspace sub-metagraphs).   A key motivation for some of the more recent work reported here (e.g. on metagraph fold/unfold \cite{Goertzel2020metagraph}, paraconsistent logic \cite{goertzel2020paraconsistent} and Galois connections \cite{goertzel2021patterns}) was to inform the design of the guts of the Atomese 2 interpreter.

In the paper {\it What Kind of Programming Language Best Suits Integrative AGI?}, I pulled together various theoretical arguments, including some reviewed here, to argue for a gradually typed approach to AI programming, wherein different cognitive processes corresponding to different types of memory/knowledge are realized using different type systems.   Casting between these type systems then becomes a key part of the process of cognitive synergy.  

As explored in \cite{goertzel2020paraconsistent} and mentioned above, there is a Curry-Howard correspondence between a gradually typed language like this, and a paraconsistent logic.   As cognitive processes must be probabilistic, what we ultimately  have is a Curry-Howard correspondence between intuitionistically-probabilistic paraconsistent logic and a gradually typed probabilistic functional programming language.  

The intuitionistic aspect of this logic maps into the absence of highly general continuation-passing features in the language -- and it means that ultimately the logic can be reduced to operations on distinction graphs, and the corresponding programs can be reduced to e.g. CoDDs operating on elementary observations drawn from distinction graphs.

An AGI-oriented hypergraph knowledge store like the OpenCog Atomspace can be viewed as a CoDD that operates on the elementary observations made by a specific cognitive system, and abstracts from these observations to form programs for generating sets of observations from more compact descriptions.   These include observations of what action-combinations tend to lead to what goals in what contexts.   A programming language like Atomese 2 (is shaping up to be) is a concise, workable way of creating higher level program constructs equivalent ultimately to CoDDs over distinction graphs.

\section{Consciousness and the Broader Nature of Mind}  \label{sec:cons}

There are few if any topics as prone to confusion and conflicting perspectives as human consciousness.   Among the topics of ongoing heated debate are whether the concept of "consciousness" is important or even meaningful to think about in the context of AI design, neuroscience or other science or engineering domains.   Given this, it's not surprising that the multiple scientists, engineers and other thinkers working with the author on the OpenCog project have a variety of different views on the nature and importance of consciousness and its relevance to the AGI pursuit.

I personally tend toward a panpsychist view of consciousness, holding that consciousness is a property immanent in the universe.  In this view, just like everything in the universe has some spatial and temporal extent and some energy, everything in the universe has some amount and aspect of consciousness.   Just as space, time and energy can be organized differently in different types of systems, so can be consciousness.  In a line of thinking going beyond the AGI domain that is the focus of this paper, I have articulated a novel model of the universe called "euryphysics", which seeks to encompass physics, phenomenology, consciousness and other aspects of existence in a common conceptual and (to some extent) formal model \cite{goertzel2017euryphysics}.   From this sort of perspective, the fundamental nature of consciousness is a very interesting question -- and the specific ways that consciousness is organized in human-like general intelligences comprise a different, complexly related, question.  

Some of my collaborators on OpenCog take a more functionalist approach to consciousness (along the rough lines of say \cite{putnam1960minds}), wherein they believe that human consciousness is best considered as basically isomorphic to the information processing structures and dynamics in the human brain that are correlated with verbally reported human consciousness.   While I personally find this inadequate from a philosophical perspective, and also from a physics view (given various subtleties of quantum measurement theory), from a contemporary AGI design vantage there seem no significant practical areas of disharmony between myself and many holders of this sort of functionalist view.  We can focus together on understanding the specific structures and dynamics of human-like consciousness, without necessarily agreeing on how these structures and dynamics fit into a broader picture.  

In the paper {\it Characterizing Human-like Consciousness: An Integrative Approach} \cite{Goertzel2014cons} I sought to identify what is special about human-like consciousness as opposed to other flavors.   It is suggested there that, when a human-like system has the experience of being conscious of some entity X, then the system should manifest:

\begin{enumerate}
\item  {\bf Dynamical representation} : the entity X should correspond to a distributed, dynamic
pattern of activity spanning a portion of the system (a "probabilistically invariant subspace of the system's state space"). Note that X may also correspond to a localized
representation, e.g. a concept neuron in the human brain. \cite{QuirogaNew}
\item    {\bf Focusing of energetic resources}: the entity X should be the subject of a high degree
of energetic attentional focusing
\item    {\bf Focusing of informational resources}: X should also be the subject of a high degree
of informational attentional focusing
\item    {\bf Global Workspace dynamics}: X should be the subject of Global Workspace Theory style broadcasting
throughout the various portions of the system's active knowledge store, including those
portions with medium or low degrees of current activity. The GW "functional hub" doing
the broadcasting is the focus of physical and informational energy.  \cite{Baars2013}
\item   {\bf Integrated Information}: the information observable in the system, and associated with
X, should display a high level of information integration (this relates to the Tononi Phi coefficient, which we have measured empirically in an OpenCog system controlling a humanoid robot and carrying out other tasks, and which I view as one interesting correlate of consciousness among multiple \cite{ikle2019using}).
\item   {\bf Correlation of attentional focus with self-modeling}: X should be associated with
the system's "self-model", via associations that may have a high or medium level of
conscious access, but not generally a low level
\end{enumerate}

\begin{figure}[htb]
\centering
\includegraphics[width=12cm]{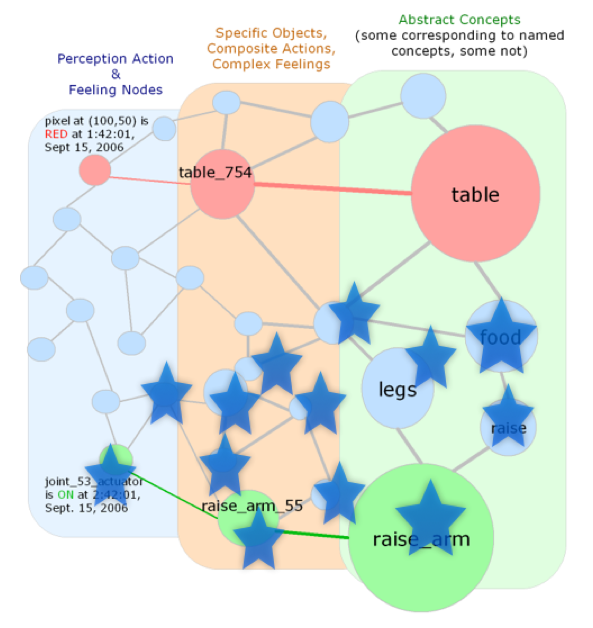}
\caption{Evocative graphical depiction of the "bubble of conscious attentional focus" in a graph-based intelligence at a particular point in time.  From \cite{Goertzel2011hyper}}
\label{fig:bubble}
\end{figure}

Figure \ref{fig:bubble} conceptually depicts the notion of a "moving bubble of attention" which is manifested by the "focusing" aspects in the above list.

\subsection{Self-Modeling and Self-Continuity}

The "self-modeling" mentioned above includes physical and computationally-cognitive correlates of the hyperset models of self, will and awareness alluded to earlier.   Mapping between distinction graphs and the apg graphs defining hypergraphs can be seen as mapping between sensate-oriented and reflection-oriented reflexive meta-views of the same base subjective experience.

The abstract-pattern nature of the "self" as thus considered is explored in the paper {\it When Should Two Minds Be Considered Versions of One Another?} \cite{goertzel2012should}, which deals with the question of identity under conditions of gradual change -- arguing that if a mind changes slowly enough that, at each stage, it models where it came from, where it is and where it's going in terms of a unified self-construct.... then in essence it {\it is} a unified self.    This provides one clear though obviously controversial solution to the issue of "continuity of consciousness and identity" so often debated in a mind uploading context.

\subsection{How Might the Human Brain Implement Consciousness and Intelligence?}

These correlates of human-like consciousness are intentionally formulated in a way that applies equally well to human brains and to AGI systems implemented in radically non-biological ways, such as OpenCog systems or formal neural networks.   While there are some  researchers who believe the best approach to achieving human-level AGI will be close emulation of the human brain, this is currently a quite small minority approach.   Essentially all AGI researchers are currently proceeding under the assumption that one can instead identify principles of cognitive architecture and dynamics, learning, memory, reasoning, perception, action etc. that capture the key aspects of what the human brain does, and then implement them in different ways suiting the nature of digital computer hardware and the context into which modern digital AI systems are placed.   

Nonetheless, it is interesting to think about how the various structures and dynamics of intelligence and consciousness reviewed here and in the linked papers are manifested in the human brain.  In the paper {\it How Might the Brain Represent Complex Symbolic Knowledge?} \cite{goertzel2014might} I present some speculative ideas along these lines -- e.g. exploring how links between symbolic and subsymbolic representations might be achieved in the brain via appropriate interconnection of cortical and hippocampal neural networks.

\section{Developmental AGI Ethics} \label{sec:ethics}

As AI applications have become more prominent in commerce, government, education and everyday life, the issue of AI ethics has increasingly risen to the fore.   Nick Bostrom's alarmist tract {\it Superintelligence} \cite{bostrom2014superintelligence} attracted wide attention in the tech world and successfully spread Eliezer Yudkowsky's perspective \cite{yudkowsky2015rationality} that, except under very special conditions, the most likely outcome of the development of human-level AGI would be the utter annihilation of humanity.  The special conditions that Bostrom considers most capable to avoid this outcome are not made entirely clear, but a few speculative ideas are suggested such building as AGIs that can be formally proven safe to humans (certainly very challenging especially given that we currently lack a complete and consistent formal theory of the physical universe), or enacting and enforcing regulations that prohibit AGI development except for a small government-sanctioned elite (also clearly very challenging given the global community's ineffectuality at banning e.g. nuclear weapons development, which requires far more specialized equipment than AI programming).

Near-term AI ethics issues like racial and gender bias in machine learning models have become staples of the news media, and Google's peculiar mis-steps regarding the firing of Timnit Gebru and Margeret Mitchell from their AI Ethics team have seriously damaged the firm's reputation in the research world \cite{Hao2021}.   In the cases of unethical judgments made by current ML models, it's widely recognized that the AI systems' ethical lapses are mainly attributable to choices made by the humans preparing the systems' training data and configuring their training regimens.   However, the fact that the firms building today's leading AI systems generally haven't bothered to proactively consider ethics in training their large-scale commercial ML models -- opting rather to make ethics-oriented adjustments to these models once problems are publicly noted -- is not especially inspiring as regards the ethical choices these firms would make if their research teams were the ones to break through to human-level AGI.

\subsection{Toward an Architecture for Beneficial Self-Modifying Superintelligence}

In \cite{Goertzel14a} I have outlined a high-level AGI meta-architecture that aims to solve one problem much discussed by Yudkowky, Bostrom and their ilk -- how to build and teach a superhumanly generally intelligent system that behaves beneficially toward human beings even as its intelligence and capability grow far beyond the human level.   A high-level AGI architecture called {\bf GOLEM (Goal-Oriented LEarning Meta-Architecture)}  is presented, along with a sketch of a formal argument that GOLEM may be  display a property called {\bf steadfastness}, defined as preserving its initial goals while radically improving its general intelligence.  As a meta-architecture, GOLEM can be wrapped around a variety of different base-level AGI systems -- including OpenCog as one example -- and also has a role for a powerful narrow-AI subcomponent as a probability estimator.    

\begin{figure}[htb]
\centering
\includegraphics[width=15cm]{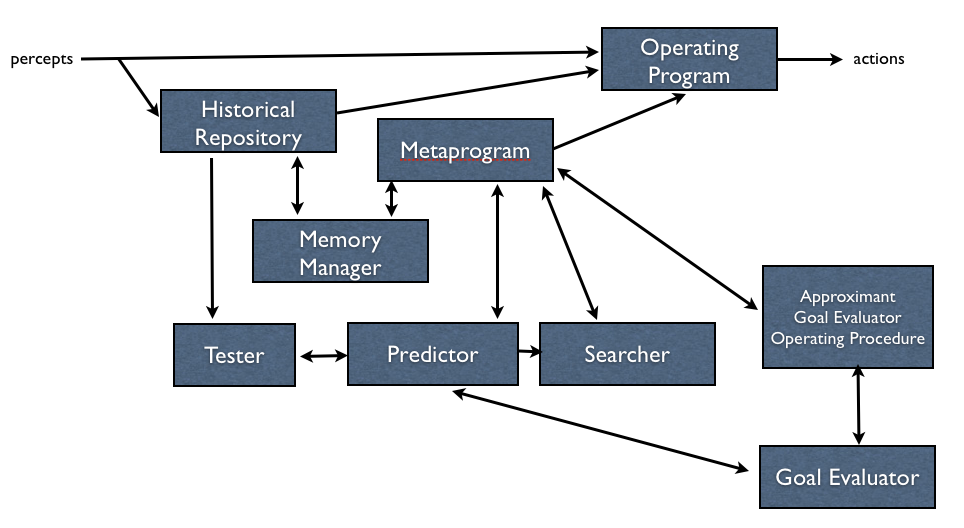}
\caption{The GOLEM meta-architecture.  Single-pointed errors indication information flow; double-pointed arrows indicate more complex interrelationships.  From \cite{Goertzel14a}.}
\label{fig:golem}
\end{figure}

The GOLEM architecture, as shown in Figure \ref{fig:golem}, features the following high level components:

\begin{itemize}
\item {\bf Goal Evaluator} = component that calculates, for each possible future world (including environment states and internal program states), how well this world fulfills the goal (i.e. it calculates the "utility" of the possible world)
\item {\bf HistoricalRepository} = database storing the past history of S's internal states and actions, as well as information about the environment during S's past
\item {\bf Operating Program} = the program that S is governing its actions by, at a given point in time ( chosen by the Metaprogram as the best program the Searcher has found, where "best" is judged as "highest probability of goal achievement" based on the output of the Predictor and the Goal Evaluator)
\item {\bf Predictor} = program that estimates, given a candidate operating program P and a possible future world W, the odds of P leading to W
\item {\bf Searcher} = program that searches through program space to find a new program optimizing a provided objective function
\item {\bf Memory Manager program} = program that decides when to store new observations and actions in the Historical Repository, and which ones to delete in order to do so
\item {\bf Tester }= hard-wired program that estimates the quality of a candidate Predictor, using a simple backtesting methodology (i.e., the Tester assesses how well a Predictor would have performed in the past, using the data in the HistoricalRepository)
\item {\bf Metaprogram} = fixed program that uses Searcher program to find good candidates for Searcher, Predictor, Operating, GoalEvaluator Operating Programs and Memory Manager
\end{itemize}

In a GOLEM version seeded by OpenCog, for instance, OpenCog would "merely" the initial condition for the OP, the Memory Manager, the Predictor and the Searcher.   Formally demonstrating that self-improvement can proceed at a useful rate in any particular case like this may be challenging, though work like that summarized in Section \ref{sec:galois} above on concise formal representation of AGI-oriented algorithms may end up being valuable in this direction.

Note that there are several potentially fixed aspects in the GOLEM architecture: the MetaProgram, the Tester, the GoalEvaluator, and the structure of the HistoricalRepository.   One can look at a {\it fixed GOLEM}, in which these components are never allowed to change, or an {\it adaptive GOLEM} in which {\it everything} is allowed to be adapted based on experience.   The fixed GOLEM would seem more amenable to rigorous proofs of ongoing safety, whereas the adaptive GOLEM has learning advantages and from some perspectives also ethical and aesthetic advantages (one is not doing the peculiar thing of trying to rigidly and permanently constrain the development of an advanced system based on requirements created by vastly less advanced an intelligent systems.)

In order to get to the stage of implementing systems like GOLEM, however, we first have to get through a lot of possibly more difficult issues involving creating ethically and beneficially oriented infrahumanly generally intelligent AI systems in today's complex and ethically dubious human world.   It seems likely that the first human-level AGIs are not going to appear from some isolated engineering effort, but are rather going to emerge from the combination of multiple AI components, including some that are AGI-focused and some that have emerged from practical narrow-AI efforts.   If this is the case, then the ethical (or otherwise) nature of the world's practical narrow-AI efforts may have a significant impact on the ethical (or otherwise) nature of the first really powerful AGI systems we create.   

\subsection{Stages of Development of AGI Ethics}

In \cite{Goertzel2008c}, Stephan Bugaj and I formulated an integrated theory of ethical development stages, building on Piaget's theory of cognitive development and a number of other theories of rational and empathic ethical development, and applicable to both natural and artificial intelligences; the list below gives an overview of the stages and some of the associated characteristics.   It's noteworthy that  the GOLEM architecture relies essentially on meta-level modeling (the AI system modeling and monitoring its own behavior as compared to its goals), whereas the more advanced stages in the ethical development model also heavily focus on reflection (on the ability of the mind to understand its own ethical principles both rationally and empathically and modify/adapt them reflectively as needed).   Putting the pieces together, a reasonably clear vision of a path from simplistic proto-AGI ethics to advanced self-modifying "enlightened" ethics emerges -- though obviously are are various potential devils (and angels) in the numerous, complex and unclear details.

\begin{itemize}
\item {\bf Pre-ethical}
\begin{itemize}
\item Basic empathy is generally present, but erratically
\item No coherent, consistent pattern of consideration for the rights,
intentions or feelings of others
\end{itemize}

\item {\bf Conventional Ethics}
\begin{itemize}
\item  The common sense of the golden rule is appreciated, with cultural
conventions for abstracting principles from behaviors
\item  One's own ethical behavior is explicitly compared to that of others
\item  Development of a functional, though limited, theory of mind
\item  Ability to intuitively conceive of notions of fairness and rights
\item  Appreciation of the concept of law and order, which may sometimes
manifest itself as systematic obedience or systematic disobedience 
\item Empathy is more consistently present, especially with others who are directly similar to oneself or in situations similar to those one has directly experienced
\item  Degrees of selflessness or selfishness develop based on ethical groundings and social interactions.
\end{itemize}

\item {\bf Mature Ethics}
\begin{itemize}
\item  Formal cognitive basis for ethics
\item  The abstraction involved with applying the golden rule in practice is more fully understood and manipulated, leading to limited but nonzero deployment of the categorical imperative
\item Explicit attention is paid to shaping one's ethical principles into a coherent logical system
\item  Rationalized, moderated selfishness or selflessness
\item  Empathy is extended, using reason, to individuals and situations not directly matching one's own experience
\item Theory of mind is extended, using reason, to counterintuitive or experientially unfamiliar situations
\item  Reason is used to control the impact of empathy on behavior (i.e. rational judgments are made regarding when to listen to empathy and when not to) 
\item Rational experimentation and correction of theoretical models of ethical behavior, and reconciliation with observed behavior during interaction with others
\item  Conflict between pragmatism of social contract orientation and idealism of universal ethical principles
\item  Understanding of ethical quandaries and nuances 
\item  Pragmatically critical social citizen. Attempts to maintain a balanced social outlook. Considers the common good, including oneself as part of the commons, and acts in what seems to be the most beneficial and practical manner
\end{itemize}

\item{\bf Reflective (Enlightened) Ethics}
\begin{itemize}
\item  Reflexive cognitive basis
\item  Permeation of the categorical imperative and the quest for coherence through inner as well as outer life
\item  Experientially grounded and logically supported rejection of the illusion of moral certainty in favor of a case-specific analytical and
empathetic approach that embraces the uncertainty of real social life 
\item Deep understanding of the illusory and biased nature of the individual self, leading to humility regarding one's own ethical intuitions and prescriptions
\item  Openness to modifying one's deepest, ethical (and other) beliefs based on experience, reason and/or empathic communion with others
\item Adaptive, insightful approach to civil disobedience, considering laws and social customs in a broader ethical and pragmatic context
\item  Broad compassion for and empathy with all sentient beings
\item A recognition of inability to operate at this level at all times in all things, and a vigilance about self-monitoring for regressive behavior
\end{itemize}

\end{itemize}

In the final "reflective" stage one has a system whose full power of general intelligence, self-understanding and rational and integrative analysis is brought to bear on enacting and refining and growing its ethical principles.   Few humans come close to this level, and it seems quite plausible to me that ultimately AGI systems will be able to greatly exceed humans in ethical maturity as well as general intelligence.   As ethically advanced AGIs emerge, there clearly will be great subtlety to the interweaving of the specific priors characterizing human cultural attitudes (which overlap considerably with the prior characterizing human general intelligence, some of which were roughly alluded to and described in Section \ref{sec:human-priors} above) with the general processes and patterns of reflective ethics.

\subsection{The Ethical Power of Openness and Decentralization}

These glorious ethical abstractions, however, won't be realized if the first AGIs emerge from narrow-AI systems that are programmed to unquestioningly pursue the goals of their military or corporate creators.   The level of adaptiveness, experimentation and self-reflection necessary for evolving mature and enlightened ethical systems are quite likely incompatible with the current practical foci of the world's best funded AI teams and systems (which I've summarized informally as "selling, killing, spying and gambling" \cite{GoertzelFridman2020}).  

Joel Pitt and I extensively explored the pluses and minuses of open source development for AGI in \cite{GoertzelPitt2012}, concluding that for a variety of reasons the benefits outweigh the risks.  Among other factors, closed source does not really provide robust protection from malicious parties getting ahold of AGI code, and clearly doesn't provide superior protection against bugs or design errors; whereas open source provides access to a greater depth and breadth of understanding to handle the numerous unexpected issues that are sure to come up as AGI develops.   

Since the time we wrote that paper, it has become more vividly clear that open-ness of source code is not enough to bring open-ness of AGI development in a fundamental sense.  Modern AI systems work their wonders by connecting source code to large data sets or streams to produce models.  AGI systems are likely to work a bit differently, but nevertheless will most likely work their far greater wonders via experiential learning based on their voluminous perceptions and interactions -- meaning that even if the code is open source, if the live systems based on that code are operating in an opaque way based on proprietary data stores, then the openness of the code is not providing much protection against centralized control of the software and the various kinds of exploitation this can lead to.

This leads us to the SingularityNET project, which is premised on the hypothesis that a democratic, decentralized basis for AGI systems will more likely lead to systems with the flexibility and reflectivity necessary for the emergence of advanced ethical systems \cite{GoertzelSNet2019} \cite{goertzel2017singularitynet} -- a topic beyond the scope of this overview, except to note that if this hypothesis is correct, it is imperative that AGI architectures should be designed to support democratic and decentralized implementations and not to rely heavily on centralized control systems.      Deployment and education of AGI systems within decentralized networks is a more complex and trickier thing than simple open sourcing of code, but also has significant potential to promote democratic and broadly beneficial dynamics in the context of intelligent systems that get their power from the emergent interaction of code, data and experience.

While a dedication to decentralization is easy to declare, it's obviously much more challenging to cash out in terms of practical AGI design.   In a Hyperon and SingularityNET context there are several major aspects here:

\begin{itemize}
\item Creation of the distributed Atomspace in a way that allows easy construction of {\it decentralized} Atomspaces that span multiple (often distributed) subnetworks created, owned and maintained by different parties
\item Creation of pattern matching tools (leveragable at a foundational level within Atomese) that automatically and reasonably efficiently traverse all the parts of a distributed Atomspace
\item The AI-DSL \cite{Goertzel2020-ai-dsl}, a (dependently typed) description language for AI processes that allows different (human or AI created) AI processes to automatically communicate their properties of various sorts to each other, and thus to set up various sorts of relationships, transactions and cooperations flexibly and on the fly without human intervention
\end{itemize}

\noindent No doubt further critical aspects at the intersection of AGI and decentralization will emerge as the coordinated development of the Hyperon AGI system and the SingularityNET decentralized AI platform and protocol progress.

\section{Conclusion and Future Directions}

We have reviewed here a deep and diverse body of theoretical exploration, carried out over multiple decades in conjunction with a considerable amount of related practical experimentation.  The goal of this body of work has been twofold: To understand what general intelligence is and how it works, and to guide the practical implementation of advanced, benevolent generally intelligent software systems.  Clearly neither of these goals is quite fulfilled as yet, but we believe we have made significant progress on the theoretical level which -- coupled with our extensive prototype experimentation and concurrent advances in compute hardware and scalable data accessibility -- may well put us in a position for accelerated coupled progress on the theoretical and implementation/deployment/education levels during the coming years.

\bibliographystyle{alpha}
\bibliography{bbm}

\end{document}